\documentclass{article}
\usepackage{arxiv}

\usepackage[utf8]{inputenc} 
\usepackage[T1]{fontenc}    
\usepackage{hyperref}       
\usepackage{url}            
\usepackage{booktabs}       
\usepackage{amsfonts}       
\usepackage{nicefrac}       
\usepackage{microtype}      
\usepackage{lipsum}		
\usepackage{graphicx}
\usepackage{natbib}
\usepackage{doi}
\usepackage{cleveref}
\usepackage{natbib}
\usepackage{makecell}

\title{Emergence of Phonemic, Syntactic, and Semantic Representations in Artificial Neural Networks}

\author{
Pierre Orhan \\
Paris Brain Institute\\
75013 Paris \\
\texttt{pierre.orhan@icm-institute.org} \\
\AND 
Pablo Diego-Simón, Emmnauel Chemla, Yair Lakretz  \\
Laboratoire de Sciences Cognitives et Psycholinguistique (LSCP)\\
Département d’Etudes Cognitives\\
École Normale Supérieure\\
PSL University, CNRS\\
75005 Paris \\
\And
Yves Boubenec \\
Laboratoire des Systèmes Perceptifs\\
Département d’Etudes Cognitives\\
École Normale Supérieure\\
PSL University, CNRS\\
75005 Paris \\
\And
Jean-Rémi King\\
Meta AI \\
}

\renewcommand{\headeright}{Preprint}
\renewcommand{\undertitle}{Preprint}
\renewcommand{\shorttitle}{Emergence of Linguistic Representations in ANNs}

\hypersetup{
pdftitle={A template for the arxiv style},
pdfsubject={q-bio.NC, q-bio.QM},
pdfauthor={David S.~Hippocampus, Elias D.~Striatum},
pdfkeywords={First keyword, Second keyword, More},
}

\begin{document}

\maketitle

\begin{abstract}
During language acquisition, children successively learn to categorize phonemes, identify words, and combine them with syntax to form new meaning.
While the development of this behavior is well characterized, we still lack a unifying computational framework to explain its underlying neural representations.
Here, we investigate whether and when phonemic, lexical and syntactic representations emerge in the activations of artificial neural networks during their training.
Our results show that both speech- and text-based models follow a sequence of learning stages: during training, their neural activations successively build subspaces, where the geometry of the neural activations represents phonemic, lexical, and syntactic structure.
While this developmental trajectory qualitatively relates to children's, it is quantitatively different: These algorithms indeed require 2 to 4 orders of magnitude more data for these neural representations to emerge.
Together, these results show conditions under which major stages of language acquisition spontaneously emerge, and hence delineate a promising path to understand the computations underpinning language acquisition.


\end{abstract}

\section{Introduction}
\begin{figure}[t]
  \centering
  \includegraphics[width=0.99\textwidth]{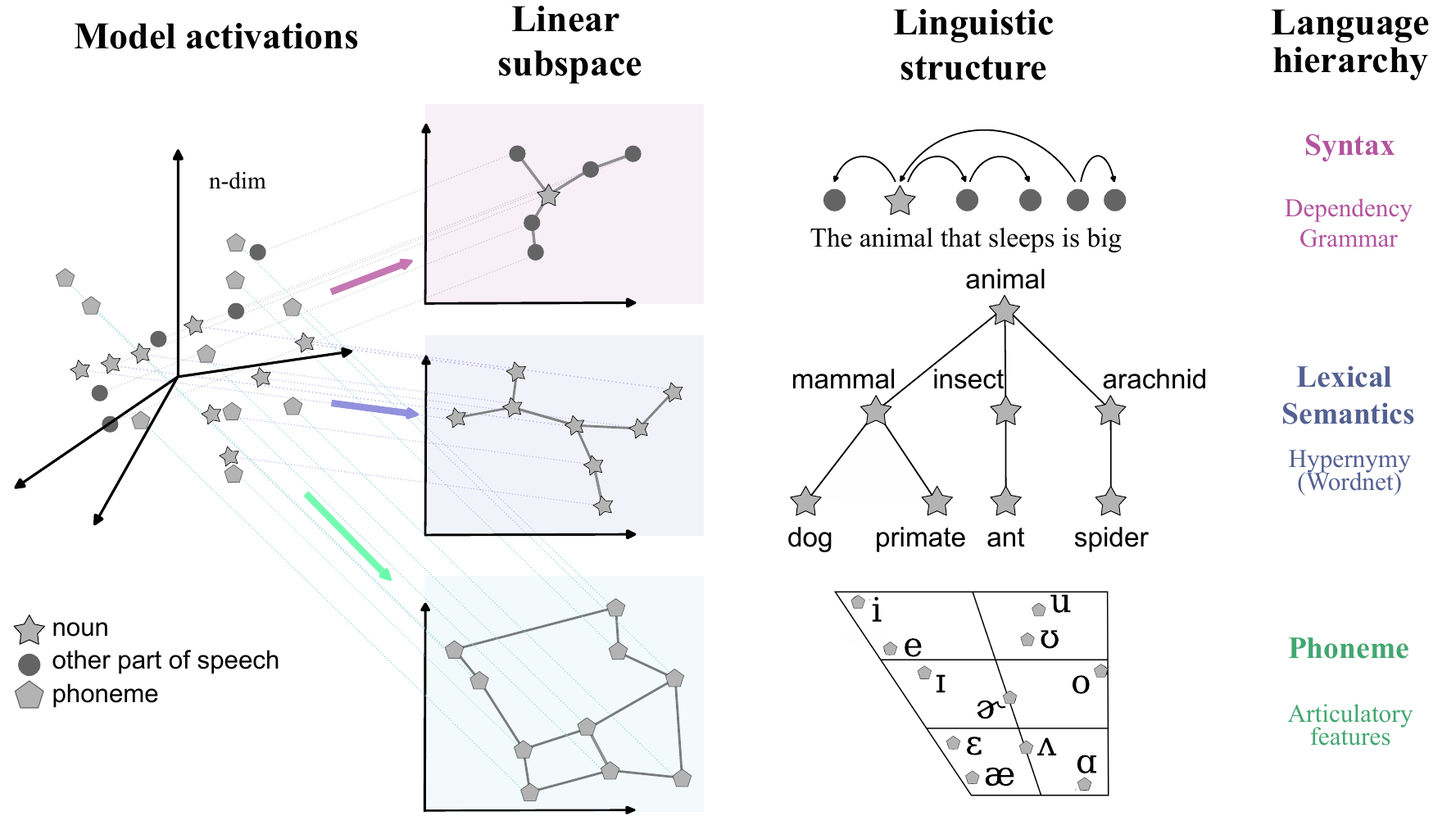}
  \caption{\textbf{Linear probing recovers phonemic, lexical semantic, and syntactic structure from the activation spaces of neural networks trained via Self-Supervised Learning.} The networks' neural activations live on a high-dimensional space in which different linear subspaces coexist. Different linear probes recover linear subspaces which represent phonemic, lexical semantic and syntactic structures, as long postulated by linguistic theories.}
  \label{fig:conceptual}
\end{figure}

To acquire language, humans follow a remarkably stereotyped developmental trajectory: infants first learn to distinguish phonological categories \citep{Kohl1993,Eilers1979,Jusczyk1993,kuhl_early_2004}, and only later become able to understand and produce complex semantic and syntactic structures \citep{Gleitman1990, Pinker1996, Christophe2008,Gomez1999, Friedmann2015, Boudreault2006}.
Linguistic theory has long posited the existence of distinct structures for these various domains of language -- for example, nested-trees for syntax, hierarchical graphs for lexical semantics, and the vowel trapezoid for phonemes. 
While the acquisition of such structures is increasingly well characterized \citep{friedmann2021growing, dupoux_cognitive_2018}, the underlying neural and computational mechanisms of language acquisition remain poorly understood \citep{kuhl2010brain,frank2011computational,martin2022plasticity,kachergis2022toward,dubois2016exploring,evanson2025emergence}.

Correlates of these structures were found to organize large language models (LLMs) activations, and consequently, LLMs appeared as a fruitful path to propose a mechanistic theory of language acquisition \citep{frank2023large,vong2024grounded,dupoux_cognitive_2018}.
Specifically, the activation patterns of self-supervised models represent simple semantic relationships \citep{landauer1997solution, Jawahar2019,Pasad2021,Pasad2024, park2025the}, syntactic structure \citep{hewitt_structural_2019, Tenney2019, evanson_language_2023} and phonological categories \citep{baevski_wav2vec_2020, de_heer_kloots_human-like_2024}.

However, it is unknown whether modern LLMs learn to instantiate different linguistic structures using a common principle.
We suggest that this gap of knowledge is due (1) to the opacity of neural networks whose representations instantiate simultaneously all linguistic structures, and (2) to the lack of longitudinal measurement of these representations during training. To fill this gap, we study the geometry of these structures in neural space and the order in which they emerge in the models during training, much like in human language acquisition. 
We generalize \citep{hewitt_structural_2019}'s structural probe to test whether and when the representations of phonemes, lexical semantics and syntactic trees emerge during the training of speech and text models.
More precisely, we test the hypothesis that these linguistic structures are neurally encoded using a geometric code. 
We evaluate whether there exists a subspace of activations in which distances between model activations linearly correspond to the phonemic, lexico-semantic, or syntactic distances predicted by linguistics.
\section{Methods}


\subsection{A Shared Linear Probe for extracting Linguistic Structures from Model Activations.}

\paragraph{Problem.}
Linguistics predicts that language inputs (e.g. text, speech) should be represented with different levels of representations (e.g. phoneme, words, syntax). 
Each level of representation can be formalized as a metric system, which indicates e.g. whether different phonemes are more or less similar to one-another or whether different words in a sentence are more or less syntactically related.
Formally, this prediction implies that language representations reduce down to a distance matrix.
Here, we aim to evaluate whether language models learn to build subspaces in which activation pattern distances correlate with these distance matrices.
For this, we extend the use of the \textit{Structural Probe} \citep{hewitt_structural_2019}, originally restricted to syntax, to any language representation, and fit this linear probe to identify subspaces that represent phonemic, lexical semantic, and syntactic features. 
Following this established interpretability paradigm, we restrict our focus to probing of representations can be directly used by downstream neurons through a simple linear readout. 
In contrast, non-linear probing could recover the target distances from arbitrary representations, thus hindering interpretability.

\paragraph{Objective.}

Consider a set of activation vectors of dimension $k$ in response to $n$ elements (e.g. word, token, phonemes): $h\in\mathbb{R}^{n,k}$.
Given a pairs of elements $(i,j)$ in the set of possible pairs $S$, and a target distance $d_{i,j}\in\mathbb{R}^+$, the \textit{Structural Probe} aims to find a linear transformation $B\in \mathbb{R}^{k,p}$ to extract the subspace where the distance between the elements' activations predict their target distance \citep{hewitt_structural_2019}.
\[
\hat{B} = argmin_{B\in \mathbb{R}^{k,p}} \sum_{(i,j)\in S}|||(h_i-h_j)B||_2^2-d_{i,k}|
\]
Following \citet{hewitt_structural_2019}, we use the squared Euclidean $||\cdot-\cdot||_2^2$ distance in the projected space.
We optimize 2D ($B\in \mathbb{R}^{k,2}$) probe for visualization and 200D ($B\in\mathbb{R}^{k,200}$) probe for evaluation.


\paragraph{Evaluation.}

For consistency with previous works \cite{hewitt_structural_2019,diego_simon_polar_2024,Limisiewicz2021}, we evaluate a probe on each level of representation (Phoneme, Lexical and Syntax) using a Spearman correlation between true distances (proposed by linguistics) and those measured in the probe space.

\subsection{Probing Datasets}

We construct a paired text-speech benchmark made of different text datasets and their artificially synthesized speech counterparts (\Cref{tab:data} in appendix).




\paragraph{Syntactic representations.}

The Universal Dependencies English Web Tree Bank (UD--EWT) corpus \citep{Silveira2014} contains 16,622 sentences extracted from different web media.
For each sentence, the syntactic tree is manually annotated by linguists according to the Universal Dependencies formalism \citep{Marneffe2021, Nivre2016}, a type of Dependency Grammar \citep{Tesniere1959}.
The UD--EWT corpus has been widely adopted to train both syntactic probes \citep{MullerEberstein2022, Diego-Simon2024, Limisiewicz2021} and syntactic parsers \citep{Grunewald2021, Hershcovich2018, Yuan2019}.
For consistency, we keep the original train, validation and test splits provided in UD--EWT.
The golden distance between two words of a sentence was defined as the number of edges between each word in the dependency tree of this sentence.

\paragraph{Lexical representations.}

Lexical semantics is assessed as a hypernymy relationship (is a generic term of) between units of senses (synsets) associated to nouns, in the WordNet semantic graph. 
For example, the synset ``animal.n.01'' is a hypernym to ``cat.n.01''. 
To obtain a maximal coverage of WordNet nouns, we construct a dedicated word-level dataset, named \emph{WordNet--Nouns}, that contains most noun synsets in WordNet, while avoiding major polysemic issues. 
This procedure is detailed in the appendix.
We focused on the noun subgraph of WordNet because including several parts of speech (POS) categories can lead to a large semantic score just because of an ability to discriminate between POS, and so, mislead to the conclusion that probes have learnt semantic structure.
This resulted in a total of 35,352 sets of words. 
For the speech modality, we construct \emph{WordNet--Nouns--TTS} by synthesizing every word with \texttt{seamless-m4t-v2-large} \citep{Barrault2025} and aligning the resulting waveforms with the Montreal Forced Aligner \citep{Mcauliffe2017}.
To define a semantic distance, we turned to path length along the WordNet gaph.
The gold distance between synsets A and B counts the number of WordNet edges (hypernym/hyponym relations) linking one synset to another. 

\paragraph{Phonemic representations.}

To create a speech counterpart to UD--EWT, we synthesize all the sentences in the UD--EWT corpus using the Text-To-Speech model \texttt{seamless-m4t-v2-large} \citep{Barrault2025}.
We then perform forced alignment with the Montreal Forced Aligner \citep{Mcauliffe2017} to obtain word and phone-level time-stamps for every synthesized utterance.
Phoneme symbols were converted from ARPABET to IPA using the panphon library \citep{mortensen-etal-2016-panphon}.
Consequently, the resulting speech corpus preserves the original UD--EWT syntactic annotations while enriching them with phoneme identity labels.
To define the distances among the different English vowel phonemes,  we used dissimilarities of articulatory feature vectors from \cite{mortensen-etal-2016-panphon}.
The dissimilarity value between 2 phonemes is an integer ranging from 0 (most similar) to 12 (most different), which represents the number of non-shared articulatory features.

\subsection{Models}


\paragraph{Text models.}
We use large language models from the LLama2 and Pythia suite \citep{Biderman2023}. Pythia models were auto-regressively trained on English text \citep{Gao2020}, differing only in their parameter size. Within this suite, we use models ranging in size from 70M to 1.4B parameters, for which the intermediate training checkpoints are publicly available \footnote{https://huggingface.co/models?other=pythia}.
The two Llama 2 models, of size 7B and 13B, allowed us to score models with stronger linguistic scores \cite{waldis_holmes_2024}, defining an upper-bound for what the smaller Pythia models could achieve.
\paragraph{Speech models.}
We use audio Wav2Vec 2.0 models  \citep{Baevski2020} previously trained on English speech, French speech, environmental sounds, and music \citep{orhan_detection_2025, Parcollet2024}. 
Wav2Vec 2.0 models are pretrained in a self-supervised fashion to reconstruct masked segments of their own latent speech representations, an approach that encourages context‑aware acoustic features. 
However, it remains unclear whether this self‑supervised objective also fosters higher‑level syntactic and semantic capabilities.









\section{Results}
\subsection{Structural probes recover the geometry of phonemic articulation.}
Phonemes can be discriminated from the activations of self-supervised speech models \citep{Baevski2020,ji_predicting_2022,seyssel_probing_2022,de_heer_kloots_human-like_2024}.
It remains unclear if their code for phonemes is structured according to articulatory features.
To measure this, structural probes are trained to identify a phoneme subspace that instantiates articulatory features.
%

\paragraph{Visualization and scores across models' layers}

\begin{figure}[t]
  \centering
  \includegraphics[width=0.99\textwidth]{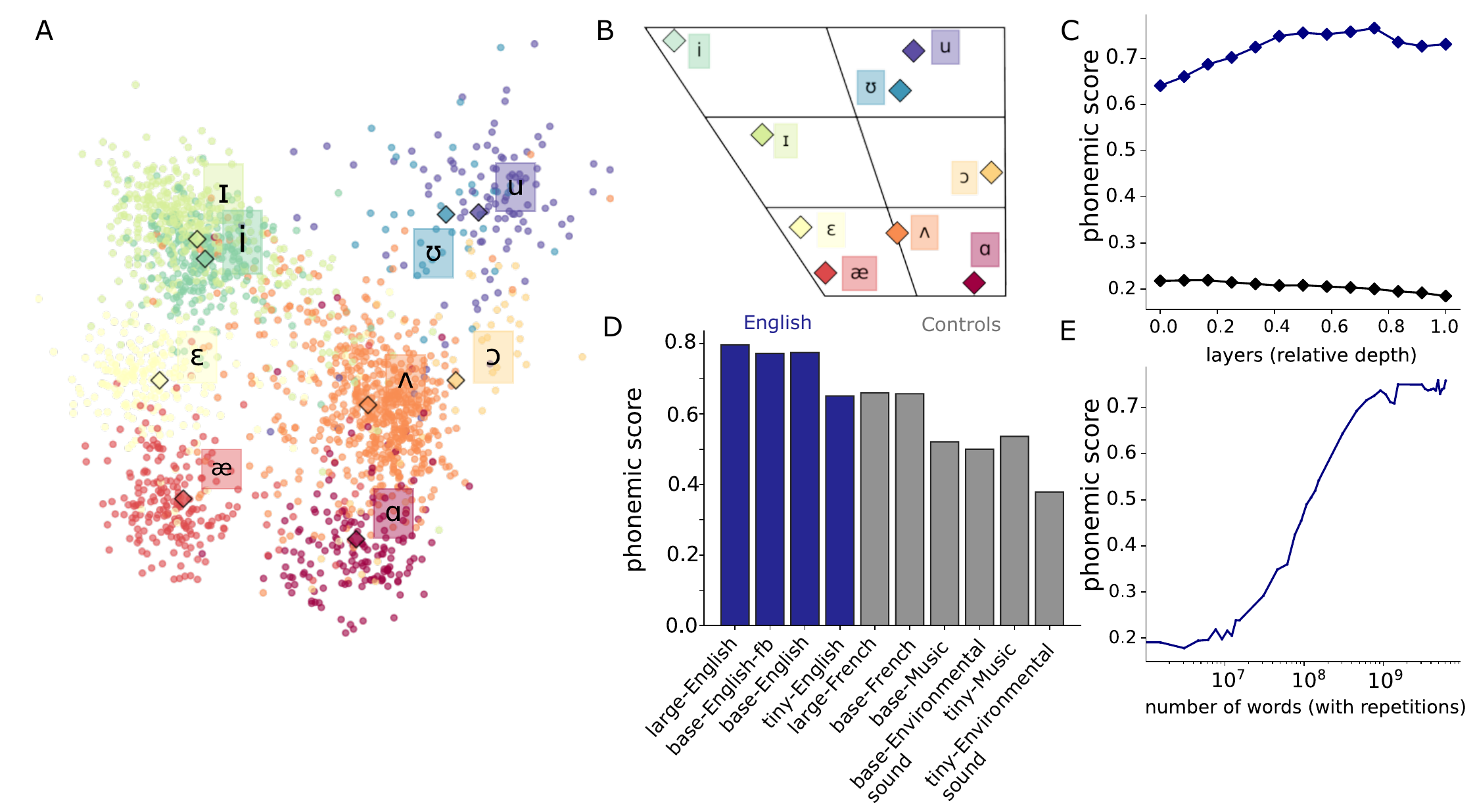}
    \caption{\textbf{Phoneme: structural probes recover the geometry of phonemic articulation.}
    \textbf{A.} Phoneme representations from a 2D structural probe trained to match pairwise articulatory distances among vowels. Responses to the same phoneme are colored identically.
    \textbf{B.} Canonical vowel trapezium indicating tongue height (vertical) and backness (horizontal).
    \textbf{C.} Layer-wise phonemic score for Wav2Vec2: probes trained on the final pretraining checkpoint peak near 80\% relative depth and consistently outperform probes trained on the initial (untrained) checkpoint.
    \textbf{D.} Effect of model and data: English pretraining and larger model capacity both yield higher phonemic scores.
    \textbf{E.} Phonemic structure emerges gradually with pretraining, with scores increasing with the number of pretraining steps; the data required for Wav2Vec 2.0 to acquire robust phonemic structure far exceeds typical estimates of children’s linguistic input.}

  \label{fig:phonemic}
\end{figure}

We first visualized the representations of a Wav2Vec 2.0 model pretrained on English speech through a less powerful 2-dimensional probe, optimized on the UD-EWT-TTS dataset.
Remarkably, the projection of the test set (Fig.~\ref{fig:phonemic}A) represents vowels similarly to the classic vowel trapezium (Fig.~\ref{fig:phonemic}B): vowel phonemes with a similar articulation tend to be closer in the phonemic subspace.
To evaluate how well the model instantiates the phoneme structure, we then train probes with a higher dimensionality.
We find that phonemes are best encoded in layers with a relative depth of 0.5 and 0.8 (Fig.~\ref {fig:phonemic}C),  achieving a Spearman correlation of 0.75. 
This phonemic score outperforms by a large margin the untrained, random, baseline (phonemic score of 0.2).
This demonstrates that instantiating the phoneme structure is not performed trivially with random acoustic processing.


\paragraph{Effect of model size.}

Is this a pure effect of model training, or is it also dependent on the model size?
We find that the phonemic score scales with model size: larger models achieve higher scores than smaller ones (Fig.~\ref{fig:phonemic}D).
Among English-trained models, Wav2Vec 2.0 large (315M parameters) outperforms base (94M), which in turn outperforms tiny (30M), each models implemented with 24, 12, and 3 Transformer layers, respectively.
This highlights that, despite its apparent simplicity, extracting the phoneme structure is easier with more contextual processing layers.
\paragraph{Controls: acoustic pretraining is not enough}
To verify that the instantiation of phoneme structure was due to speech processing instead of generic acoustic processing, we compared models trained on different data.
Phonemic scores depend strongly on the training data (Fig.~\ref{fig:phonemic}D): models trained on English speech perform best, whereas those trained on French, music, or environmental sounds perform poorly. 
Note also, that the music and environmental sounds models are not completely naive to speech, with an estimated 8\% of speech inputs in the environmental sound dataset \citep{orhan_detection_2025}.
Only the French large and base models outperform the \texttt{tiny} English model, which is expected given the model’s limited size.
These controls prove that the instantiation of phoneme structure results from the emergence of speech-specific processes.
\paragraph{Emergence during pretraining}
How is this representation of phonemes learned?
We probe the English Wav2Vec 2.0 94M model across its training checkpoints. 
For simplicity, we focus on the layer with the best phonemic score, i.e., layer 9. 
Fig.~\ref{fig:phonemic}E shows a gradual rise in this score with the cumulative spoken-word count, reaching emergence after roughly $10^9$ words (including repetitions), which corresponds to about $90{,}000$ pretraining steps.
This is considerably more than the $10^7$ words that children experience in their first year on average \citep{gilkerson_mapping_2017}.
Taken together, these results indicate that self-supervised speech models spontaneously learn to instantiate a phoneme subspace whose geometry mirrors distances of articulatory features. 
\subsection{Lexical Semantics}
Language model representations have been known to encode semantic features \citep{korchinski2025emergence}.
Additionally, models' behavior and the organization of their final layer have been shown to be consistent with an instantiation of lexical semantic relationship \citep{cohen_crawling_2023}, including the WordNet graph \citep{park_linear_2024}.
To measure the internal layers' representation, we train structural probes to predict the WordNet noun graph from text and speech models.

\paragraph{Visualization and scores across models' layers. }

\begin{figure}[t]
  \centering
  \includegraphics[width=0.99\textwidth]{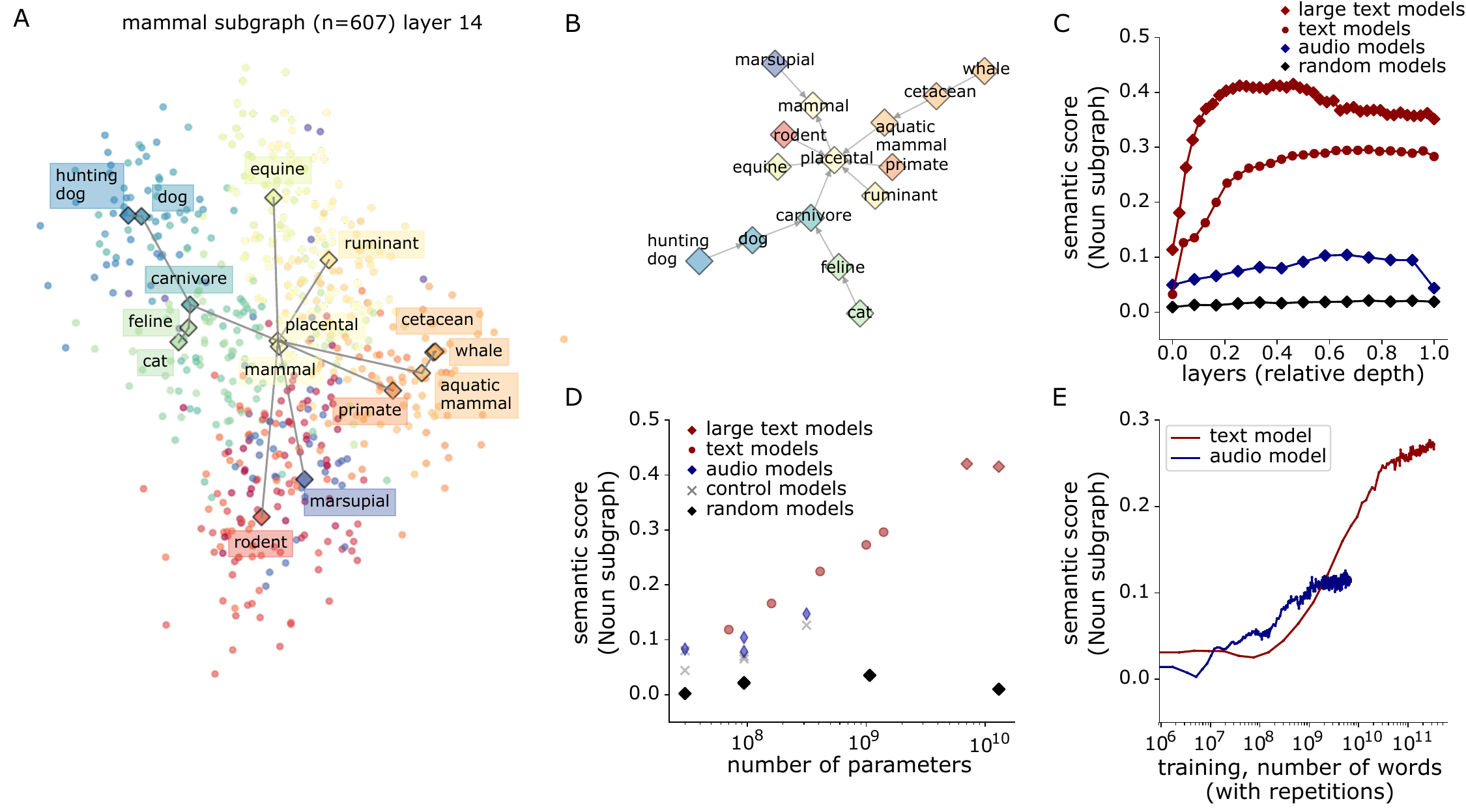}
  \caption{\textbf{Structural probes recover the geometry of lexical  semantic graphs}
  A: 2d projection of text model (pythia-1.4B) response to all words in the mammal subgraph.
  B: Example WordNet subgraph, displaying synsets of the mammal subgraph with more than 50 hyponyms.
  c: Semantic score (for the graph composed of all nouns) for a large text (Llama-13B), text (pythia-1.4B), audio (Wav2vec 2.0-94M base), and random (Wav2Vec 2.0-94M base) models.
  D: Semantic scores for all models as a function of the model size.
  E: Semantic score as a function of the quantity of pertaining for the text (pythia-1.4B) and audio (Wav2vec 2.0-94M base) models.
  }
  \label{fig:semantic}
\end{figure}

We first optimized a 2-D probe of text models' semantic representations in order to visualize their representations. 
We display the training and test elements together while evaluating only across test elements.
To our surprise, a 2-D probe already embeds well-constructed semantic graphs, especially when we focus on small subsets of the larger WordNet tree.
As an example, we plot in Fig.~\ref{fig:semantic}A, a 2-D embedding of the tree below the mammal synset from the activations of a Pythia-1.4B text model.
Other examples for several ``supersense'' semantic categories are available in Figs.~\ref{fig:sup_semantic1} and \ref{fig:sup_semantic2} in appendix, as well as an example of 2-D embedding of the whole WordNet tree, see Fig.~\ref {fig:sup_semanticWordnet} in appendix.
In two dimensions, this embedding does not capture the exact tree, but rather positions well the largest category of this semantic tree, as illustrated by plotting these categories' centroid in Fig.~\ref{fig:semantic}A-B (diamond markers).
Smaller categories and the exact edges of the tree require a larger embedding space. 
We consequently train and evaluate a 200-D probe of the whole WordNet graph, in text and audio models.
For this 200-D probe, the semantic scores across model layers follow an inverted U-shape in all models: Wav2vec 2.0 (160M, max 0.1), Pythia (1.4B, max 0.29), and Llama2 (13B, max 0.41), Fig.~\ref{fig:semantic}C.
This is remarkable, because transformer's tokens are vectorised and could therefore contain some semantic information. 
Here, we show that such semantic information is not sufficient to instantiate the lexical semantic structure of nouns. 
Instead, several computational steps are performed to merge and position these tokens' embedding together into a semantic subspace.

\paragraph{Effect of model size.}

This questions the computational power necessary for these semantic spaces to emerge.
We measured the maximal semantic score across layers as a function of parameters in a set of text and audio models, Fig.~\ref{fig:semantic}D.
Remarkably, semantic scores increase linearly with text model sizes in the Pythia model suite and finally plateau at a large (7B) size for the Llama2 models.
Similarly, the semantic scores of audio models increase with model size, although through a less steep slope.
At equivalent size, text models have slightly larger semantic scores than audio models.
This is in stark contrast with their very distinct training regime, with audio models seeing a completely distinct amount of data. 
This hints at a possible different training dynamic between audio and text models.

\paragraph{Emergence during pretraining. }

We measured the evolution of the semantic score as a function of pretraining of a base Wav2vec 2.0 model (94M) trained on librispeech data, and a 1B Pythia model, Fig.~\ref{fig:semantic}E.
Audio models' semantic scores initially increased faster than text models' scores, but then saturated.
This saturation of the audio models questions whether they discovered semantic structure or if these scores could be explained by the acoustic properties of the words.

\paragraph{Controls: semantic mostly arise from acoustic cues in audio models}

Acoustic cues could be sufficient to encode a small part of the semantic subgraph. 
For example, ``dog'' and ``hunting dog'' are expected to be close by in the graph, but this could be predicted purely from acoustic features; here, the fact that ``dog'' is present in both sets of words.
To control for these acoustic features, we evaluated tiny (30M) and base (94M) Wav2vec 2.0 models trained on music, environmental sounds, and a large (315M) model trained on French speech.
Each of these audio models reached significant semantic scores, and these scores were in the same range as the semantic scores of models exposed to English speech, but remained much lower than the scores of text models (Table \ref{tab:semantic} in appendix).
With this structural probe, it consequently seems that most of the audio models' scores are due to acoustic correlates. 
Despite their low semantic score, the English-exposed models better clustered 89\% of semantic categories of the WordNet tree than control models, with a 0.1 F1-score gain on average (Fig~\ref{fig:sup_scatter_control_semantic} in appendix, paired t-test $3.8\times10^{-36}$).
Together, these results demonstrate a clear and modest but significant instantiation of lexical semantic structure by text and audio models, respectively.

\subsection{Syntax}
Language models are known to code syntactic trees through distances \citep{hewitt_structural_2019}, while syntactic trees per se have yet to be found in speech models, since previous approach focused on decoding statistical quantities derived from the trees, like node depth \citep{pasad_what_2024}.
Here we train structural probes to measure if speech models also use a distance code to instantiate syntactic trees. 

\begin{figure}[t]
  \centering
  \includegraphics[width=0.99\textwidth]{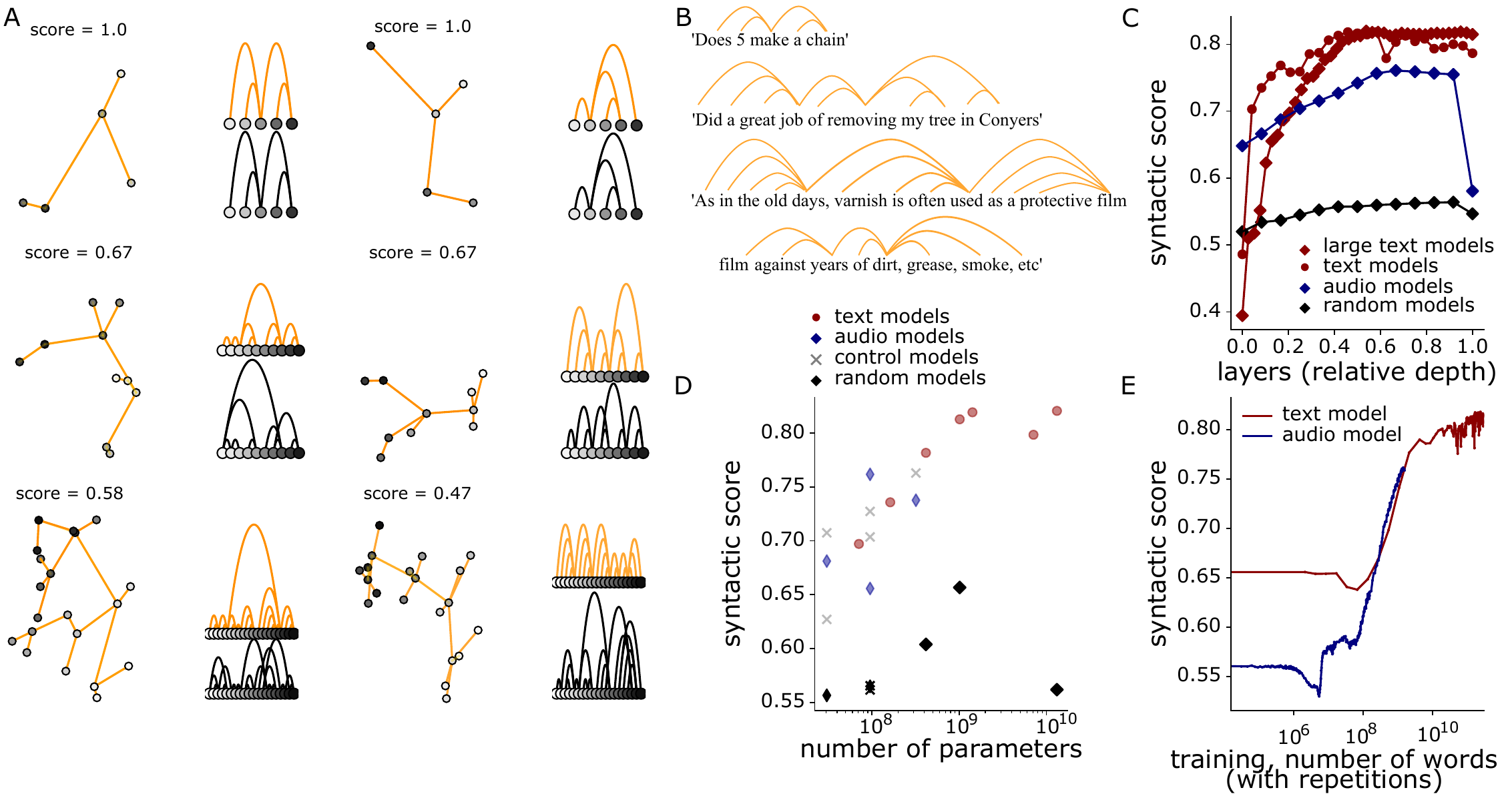}
  \caption{\textbf{Structural probes recover the geometry of syntactic trees} A. In each left subpanel, we plot an example of a 2d probe projection of the model activations, with trees reconstructed through the minimum spanning tree algorithm. On the right subpanel, we plot the gold (yellow) and predicted (black).
  B. Example of sentences and their syntactic trees.
  C. Syntactic scores of a large text (Llama-13B), a text (pythia-1B), speech (Wav2vec 2.0-94M base) and random (Wav2vec 2.0-94M base) models for each transformer layer. D. Highest syntactic score across layer for each model. E. Emergence of the syntactic abilities during pretraining, which includes several repetitions of the same dataset for the audio model.}
  \label{fig:syntactic}
\end{figure}

\paragraph{Visualization and scores across models’ layers.}
We first visualized the 2-D probing of a Wav2Vec 2.0 base (94M) model activations in response to 6 sentences with increasing hierarchical depth (Fig.~\ref{fig:syntactic}A and B).
As expected, the true (gold) and predicted (black) syntactic trees closely matched for simple sentences, but became increasingly more distant for harder, complex sentences. 
We then evaluated syntactic scores with probing in 200 dimensions.
We observed strong syntactic scores for all models, including Wav2vec 2.0 models (94M parameters, max: 0.76), Pythia model (1.4B, max: 0.81), Llama2 model (13B, max: 0.82) (Fig.~\ref{fig:syntactic}C).

\paragraph{Effect of model size}
For text and audio models, syntactic performances first increase then plateau as a function of the number of parameters, Fig.~\ref{fig:syntactic}D.
Remarkably, unlike semantic scores, text models' performances quickly plateau with model size.
This indicates that syntactic structure computations do not require very large text models, but also that large text models do not develop a perfect syntactic parsing algorithm.

\paragraph{Emergence during pretraining}
We plot in Fig.~\ref{fig:syntactic}E the evolution of the syntactic score as a function of pretraining.
Unlike text models, audio models' scores did not seem to saturate with the quantity of pretraining.
For models of equivalent size, the emergence of syntax was faster in audio models, for which the dataset also used a smaller quantity of different sentences.
This suggests that audio data provides syntactic cues that the model can discover and use to build syntactic subspaces.
 
\paragraph{Controls: models instantiating syntax encode more than a linear tree}
Similarly to semantics, could syntax processing in audio models be explained by acoustic features?
If indeed control audio models reached relatively high syntactic scores, we found that this came from the oversimplicity of the metric.
Indeed, control audio models partially predicted the syntactic trees, purely from the fact that they could predict a linear tree (where words are linked to their neighbors).
To demonstrate this, we computed the Spearman correlation with a linear tree and compared it to the correlation with the gold dependency tree.
This correlation were similar in the case of control models but not for English-exposed models.
Critically, this was not the case for English-exposed models whose projected representations correlated more with the syntactic than the linear structure.
This demonstrates that true syntactic representations in response to English sentences are present in English-exposed models but not in models trained on French, music, or environmental sounds.

\subsection{Order of acquisition}

\begin{figure}[t]
  \centering
  \includegraphics[width=0.99\textwidth]{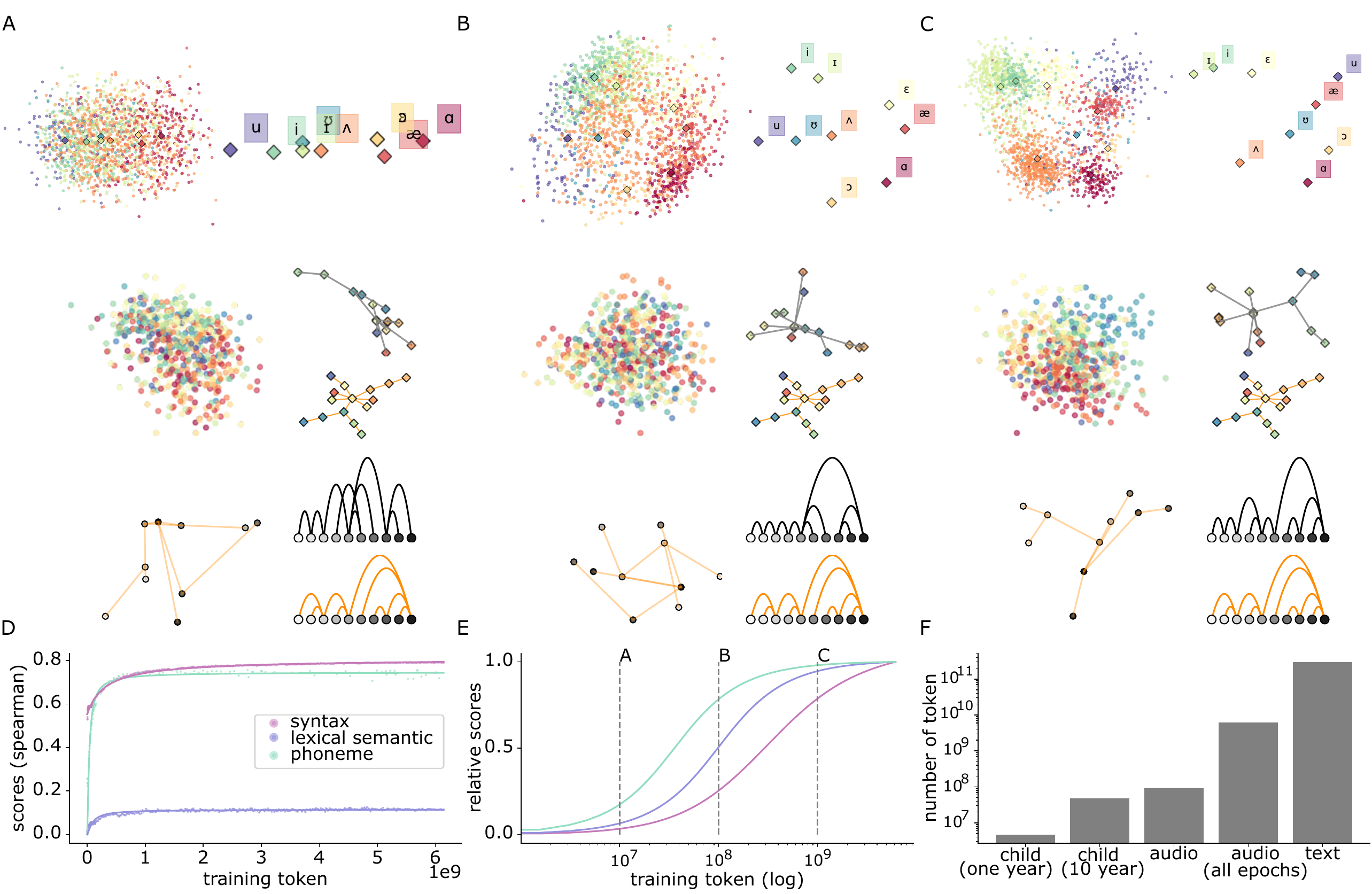}
  \caption{\textbf{Linguistic structures are acquired in sequential order}
  A, B and C: 2d visualizations of the phonemic and semantic space along with one example syntactic structure for 3 pertaining steps.
  D: Left: Emergence of semantic, syntactic, and phonemic score for a Wav2vec 2.0-94M base model (dots),  along with a parametric fit (line)
  E: relative scores of the parametric fit (min to 0 and max to 1), demonstrating the successive emergence of phonemic and lexical-syntactic structures. Checkpoints used for the top sub-panel are highlighted with lines.
  F: average amount of words (tokens) heard by a child or used by models. Note that for the audio dataset, many epochs of optimization over the same dataset take place.}
  \label{fig:emergence}
\end{figure}

We next compare the learning dynamics of these phonemic, semantic, and syntactic codes in the audio model. 
To reveal these emergences, we repeated phonemic, syntactic, and semantic probing across checkpoints logarithmically spanning the pretraining of the models.
We plot in Fig.~\ref{fig:emergence}A-C the probing of 3 checkpoints that highlight the successive emergence of phonemic and lexical-syntactic structures in the audio models.
We plot in Fig.~\ref{fig:emergence}D the emergence of these scores and fit a parametric curve for each. 
This allows us to then plot in Fig.~\ref{fig:emergence}E a relative score for each curve and compare their dynamic.
Remarkably, we observe that phonemic emergence is followed by a partial emergence of lexical semantic and finally an emergence of syntactic abilities in the audio models.
These emergences are each roughly separated by an order of magnitude of input data, indicating that each stage solidifies before the previous one.
Yet, as a semantic code only partially emerges in the audio model, it remains to be seen if these models truly predict that lexical semantic structure appears before syntactic structure.

\section{Discussion}
\paragraph{Goal and approach}
To evaluate whether modern language algorithms can help understand the computational underpinning of language acquisition, we generalized \citet{hewitt_structural_2019}'s Structural Probe and assessed the emergence of phonemic, lexical, and syntactic representations during the training audio and text models.

\paragraph{Similarities to Human Language Acquisition.}
In spite of their remarkably distinct architecture and learning principles, audio models appear, like children, to spontaneously build linguistic features, in a developmental trajectory that, at least at a coarse view, follows a phonemic and lexico-syntactic sequence.
This emergence effectively captures some of the developmental trajectory of children, who -- during their first year of life -- demonstrate a progressive categorization of phonemes, and subsequently develop an ability to understand and produce syntactic structures from an increasingly large and precise vocabulary \citep{Gomez1999,kuhl_early_2004,dupoux_cognitive_2018, Friedmann2015, Boudreault2006,sainburg2022long}. 

\paragraph{A quantitative gap.}
Consistent with previous work \citep{dupoux_cognitive_2018, lavechin_can_2022, frank2011computational, hu2024findings}, data efficiency remains a major gap: compared to human children, the presently-studied speech and text models require 2-4 orders of magnitude more input data for these representations to emerge (Fig.~\ref{fig:emergence} F) \citep{gilkerson_mapping_2017}.
This discrepancy highlights that modern language algorithms remain remarkably inefficient, and thus calls for exploring novel neural architectures and training paradigms \citep{wolf2023whisbert,hu2024findings,xiang_visualizing_2022}.
Part of this gap may be explained by non-language data pretraining. 
Indeed, \citet{poli2024modeling} showed that pretraining speech models on ambient sounds improves data-efficiency to then develop phonemic representations.
Similarly, \citet{orhan_detection_2025} showed that pretraining on ambient sounds spontaneously lead audio models to detect algebraic auditory structures.

\paragraph{Modeling the developmental trajectory of representations.}
This emergence strengthens earlier work on toy models.
For example, \citep{kemp_discovery_2008} showed that hierarchical concept structures can sequentially emerge from feature spaces by iteratively searching over a structured hypothesis space.
More formally, \citet{saxe_exact_2014, saxe_mathematical_2019} showed mathematically that hierarchical representations emerge sequentially in deep linear networks during training. 
Although we did not observe a sequential emergence of semantic categories (see Fig.~\ref{fig:knnScores} in appendix), we observe a sequential emergence across linguistic structures. 
Finally, \citet{Lavechin2025, chen2024sudden} show that self-supervised learning from raw speech and text yields a staged emergence of phonetic, early lexical structure and syntax without explicit labels.
The present work complements these approaches by testing a unified code within self-supervised neural networks for the different linguistic structures posited by linguists.

\paragraph{Limitations}
Our findings come with several limitations.
First, the developmental ordering between syntax and lexical semantics, a central question in cognitive science \citep{Gleitman1990,Pinker1994}, remains unresolved. 
Second, we report aggregate scores and do not stratify performance by simple versus complex syntactic phenomena or by token/phoneme frequency. 
Third, our evaluation relies exclusively on structural probing of internal representations, rather than on prediction-based tests at the phonemic, syntactic, or lexical levels \citep{schatz_evaluating_2013,warstadt2020,hill2015,nguyen_zero_2020}. 
Thus, our conclusions concern representation rather than inference and may not fully characterize downstream performance.

Fourth, optimizing a probe to match pairwise distances might nevertheless be too restrictive.
This approach might nevertheless be too restrictive. For example, the distance between “dog” and “river” does not have to be as precisely captured as the distance between “dog” and “cat”.
Additionally, the exact choice of distance, for example, the shortest path distance on the WordNet graph, is itself questionable.
To show that our approach can be generalized beyond distance learning, while simultaneously strengthening our results, we present supplementary results using a “contrastive” probe, which optimizes the extraction of the topology of the structure, irrespective of the distance.
This probe is optimized according to the contrastive objective of \citep{nickel_poincare_2017}.  
On measures of topological fidelity (unlabeled undirected attachment score for syntax, rank for syntax and semantic), this probe achieved better scores than the distance probe, while revealing similar emergence dynamics (Fig.~\ref{fig:sup_contrastive} in appendix).
These results prove that our approach could, in future work, be extended by searching for different probing objectives.

Fifth, we do not probe phonemic structure in text models. 
This is because text models for which checkpoints were available used token-level inputs instead of letters or grapheme inputs.

Sixth, our analysis is restricted to nouns to avoid equating semantics and part-of-speech categorization.
This analytical choice may limit the generality of the present finding. 
Future work should thus extend to other parts of speech and explore semantic relationships beyond hypernymy.


\paragraph{Impact and Future Directions}
The present generalization of \citet{hewitt_structural_2019}'s structural probe to lexical semantics and phonology provides a unifying framework to characterize the structure and emergence of linguistic representations in neural networks.
This approach reduces the notion of `representation' to (squared) Euclidean distances between elements in specific subspaces of the activation space. 
This approach also provides a mechanistic explanation: As linear readout is a core operation of any neuron, linear subspaces make these language representations readily accessible to any subsequent steps \citep{kriegeskorte_deep_2015,higgins2018towards,park_linear_2024}, including attention mechanisms.
Our results, like previous interpretability works, suggest the existence of laws governing the geometry of neural networks' activations \cite{elhage_toy_2022,hernandez_linearity_2024,jha_harnessing_2025,park2025the,landauer1997solution,huh_platonic_2024}.

Our probing approach could be used to understand how structures are built through in-context learning. 
For example, \citep{orhan_detection_2025} show that algebraic auditory structures are detected by speech models through in-context learning. 
This work predicts that the code underlying this detection is a distance code which can be detected through a linear probing approach.

Perhaps most excitingly, this work delineates a path to model the brain bases of language acquisition \citep{di2023emergence,nakagi2025triple}.
In particular, \citet{evanson2025emergence} show that the representations learned by audio and text based models not only linearly align with those of the human brain \citep{huth_natural_2016,Caucheteux2022, Millet2022,oota2023joint,goldstein2022shared}, but their developmental trajectory also captures the change of representations in the brain of children during their development. 
Overall, these findings suggest that modern language models are not only powerful tools for analyzing complex data but also playgrounds from which we can theorize and understand computational principles underlying language acquisition in the human brain.

\subsection*{Acknowledgments}

This project has received funding from the European Union’s Horizon 2020 research and innovation program under the Marie Sklodowska-Curie grant agreement No 945304 to PO and PDS. 
This work was granted access to the HPC resources of IDRIS under the allocations 2023-AD011014524, 2024-AD011014524R1 and 2025-AD011014524R2 made by GENCI (P.Orhan), and 2023-AD011014766 (PDS).
The author would like to thank IDRIS for their support on the Jean-Zay cluster.
This project received funding from PSL University under the grant agreement ANR-10-IDEX-0001-
02 (YB,YL,EC).
This project received funding from the Departement d’ ´ Etudes Cognitives (DEC) at ENS under the ´
grant agreement FrontCog, ANR-17-EURE-0017 (YB,YL,EC).
This project received funding from the French National Research Agency (ANR) under the grant
agreement ComCogMean, Projet-ANR-23-CE28-0016 (EC).
The funders had no role in study design, data collection and analysis, decision to publish, or preparation of the manuscript.

\subsection*{Reproducibility Statement}

We take the following steps to ensure reproducibility of this work:
All models necessary to replicate this study can be downloaded from the HuggingFace repository, following Table \ref{tab:model_comparison} in appendix.
All hyperparameters for every structural probe used in this study have been listed in Table \ref{tab:probing_hyperparameters} in appendix.
All dataset generation and cross-validation procedures are presented in detail in the appendix.

\bibliographystyle{iclr2026_conference}
\bibliography{iclr2026_conference}  

@article{zada_podcast_2025,
	title = {The “{Podcast}” {ECoG} dataset for modeling neural activity during natural language comprehension},
	volume = {12},
	copyright = {2025 The Author(s)},
	issn = {2052-4463},
	url = {https://www.nature.com/articles/s41597-025-05462-2},
	doi = {10.1038/s41597-025-05462-2},
	language = {en},
	number = {1},
	urldate = {2025-11-20},
	journal = {Scientific Data},
	author = {Zada, Zaid and Nastase, Samuel A. and Aubrey, Bobbi and Jalon, Itamar and Michelmann, Sebastian and Wang, Haocheng and Hasenfratz, Liat and Doyle, Werner and Friedman, Daniel and Dugan, Patricia and Melloni, Lucia and Devore, Sasha and Flinker, Adeen and Devinsky, Orrin and Goldstein, Ariel and Hasson, Uri},
	month = jul,
	year = {2025},
	note = {Publisher: Nature Publishing Group},
	pages = {1135},
}

@inproceedings{xiang_visualizing_2022,
	address = {Dublin, Ireland},
	title = {Visualizing the {Relationship} {Between} {Encoded} {Linguistic} {Information} and {Task} {Performance}},
	doi = {10.18653/v1/2022.findings-acl.35},
	urldate = {2025-11-19},
	booktitle = {Findings of the {Association} for {Computational} {Linguistics}: {ACL} 2022},
	publisher = {Association for Computational Linguistics},
	author = {Xiang, Jiannan and Li, Huayang and Lian, Defu and Huang, Guoping and Watanabe, Taro and Liu, Lemao},
	editor = {Muresan, Smaranda and Nakov, Preslav and Villavicencio, Aline},
	month = may,
	year = {2022},
	pages = {410--422},
}

@inproceedings{cohen_crawling_2023,
	address = {Dubrovnik, Croatia},
	title = {Crawling {The} {Internal} {Knowledge}-{Base} of {Language} {Models}},
	url = {https://aclanthology.org/2023.findings-eacl.139/},
	doi = {10.18653/v1/2023.findings-eacl.139},
	booktitle = {Findings of the {Association} for {Computational} {Linguistics}: {EACL} 2023},
	publisher = {Association for Computational Linguistics},
	author = {Cohen, Roi and Geva, Mor and Berant, Jonathan and Globerson, Amir},
	editor = {Vlachos, Andreas and Augenstein, Isabelle},
	month = may,
	year = {2023},
	pages = {1856--1869},
}

@misc{elhage_toy_2022,
	title = {Toy {Models} of {Superposition}},
	url = {http://arxiv.org/abs/2209.10652},
	doi = {10.48550/arXiv.2209.10652},
	urldate = {2025-09-23},
	publisher = {arXiv},
	author = {Elhage, Nelson and Hume, Tristan and Olsson, Catherine and Schiefer, Nicholas and Henighan, Tom and Kravec, Shauna and Hatfield-Dodds, Zac and Lasenby, Robert and Drain, Dawn and Chen, Carol and Grosse, Roger and McCandlish, Sam and Kaplan, Jared and Amodei, Dario and Wattenberg, Martin and Olah, Christopher},
	month = sep,
	year = {2022},
	note = {arXiv:2209.10652 [cs]},
}

@misc{hernandez_linearity_2024,
	title = {Linearity of {Relation} {Decoding} in {Transformer} {Language} {Models}},
	url = {http://arxiv.org/abs/2308.09124},
	doi = {10.48550/arXiv.2308.09124},
	publisher = {arXiv},
	author = {Hernandez, Evan and Sharma, Arnab Sen and Haklay, Tal and Meng, Kevin and Wattenberg, Martin and Andreas, Jacob and Belinkov, Yonatan and Bau, David},
	month = feb,
	year = {2024},
	note = {arXiv:2308.09124 [cs]},
}

@misc{park_linear_2024,
	title = {The {Linear} {Representation} {Hypothesis} and the {Geometry} of {Large} {Language} {Models}},
	url = {http://arxiv.org/abs/2311.03658},
	doi = {10.48550/arXiv.2311.03658},
	urldate = {2025-09-23},
	publisher = {arXiv},
	author = {Park, Kiho and Choe, Yo Joong and Veitch, Victor},
	month = jul,
	year = {2024},
	note = {arXiv:2311.03658 [cs]},
}

@inproceedings{baevski_wav2vec_2020,
author = {Baevski, Alexei and Zhou, Henry and Mohamed, Abdelrahman and Auli, Michael},
title = {wav2vec 2.0: a framework for self-supervised learning of speech representations},
year = {2020},
isbn = {9781713829546},
publisher = {Curran Associates Inc.},
address = {Red Hook, NY, USA},
booktitle = {Proceedings of the 34th International Conference on Neural Information Processing Systems},
articleno = {1044},
numpages = {12},
location = {Vancouver, BC, Canada},
series = {NIPS '20}
}

@incollection{Pinker1994,
    author = {Steven Pinker},
    isbn = {9780262286688},
    title = {How could a child use verb syntax to learn verb semantics ?},
    booktitle = {The Acquisition of the Lexicon},
    publisher = {The MIT Press},
    year = {1994},
    month = {10},
    doi = {10.7551/mitpress/1075.003.0017},
    url = {https://doi.org/10.7551/mitpress/1075.003.0017},
    eprint = {https://direct.mit.edu/book/chapter-pdf/2437976/9780262286688_caq.pdf},
}

@article{hill2015,
    title = "{S}im{L}ex-999: Evaluating Semantic Models With (Genuine) Similarity Estimation",
    author = "Hill, Felix  and
      Reichart, Roi  and
      Korhonen, Anna",
    journal = "Computational Linguistics",
    volume = "41",
    number = "4",
    month = dec,
    year = "2015",
    address = "Cambridge, MA",
    publisher = "MIT Press",
    url = "https://aclanthology.org/J15-4004/",
    doi = "10.1162/COLI_a_00237",
    pages = "665--695"
}

@article{warstadt2020,
    title = "{BL}i{MP}: The Benchmark of Linguistic Minimal Pairs for {E}nglish",
    author = "Warstadt, Alex  and
      Parrish, Alicia  and
      Liu, Haokun  and
      Mohananey, Anhad  and
      Peng, Wei  and
      Wang, Sheng-Fu  and
      Bowman, Samuel R.",
    editor = "Johnson, Mark  and
      Roark, Brian  and
      Nenkova, Ani",
    journal = "Transactions of the Association for Computational Linguistics",
    volume = "8",
    year = "2020",
    address = "Cambridge, MA",
    publisher = "MIT Press",
    url = "https://aclanthology.org/2020.tacl-1.25/",
    doi = "10.1162/tacl_a_00321",
    pages = "377--392"
}

@misc{saxe_exact_2014,
	title = {Exact solutions to the nonlinear dynamics of learning in deep linear neural networks},
	url = {http://arxiv.org/abs/1312.6120},
	doi = {10.48550/arXiv.1312.6120},
    year = {2014},
	number = {{arXiv}:1312.6120},
	publisher = {{arXiv}},
	author = {Saxe, Andrew M. and {McClelland}, James L. and Ganguli, Surya},
	urldate = {2022-07-29},
	date = {2014-02-19},
	eprinttype = {arxiv},
	eprint = {1312.6120 [cond-mat, q-bio, stat]},
}

@article{sainburg2022long,
  title={Long-range sequential dependencies precede complex syntactic production in language acquisition},
  author={Sainburg, Tim and Mai, Anna and Gentner, Timothy Q},
  journal={Proceedings of the Royal Society B},
  volume={289},
  number={1970},
  pages={20212657},
  year={2022},
  publisher={The Royal Society}
}

@article{wolf2023whisbert,
  title={WhisBERT: Multimodal Text-Audio Language Modeling on 100M Words},
  author={Wolf, Lukas and Tuckute, Greta and Kotar, Klemen and Hosseini, Eghbal and Regev, Tamar and Wilcox, Ethan and Warstadt, Alex},
  journal={arXiv preprint arXiv:2312.02931},
  year={2023}
}

@article{di2023emergence,
  title={Emergence of the cortical encoding of phonetic features in the first year of life},
  author={Di Liberto, Giovanni M and Attaheri, Adam and Cantisani, Giorgia and Reilly, Richard B and N{\'\i} Choisdealbha, {\'A}ine and Rocha, Sinead and Brusini, Perrine and Goswami, Usha},
  journal={Nature communications},
  volume={14},
  number={1},
  pages={7789},
  year={2023},
  publisher={Nature Publishing Group UK London}
}

@inproceedings{schatz_evaluating_2013,
	title = {Evaluating speech features with the minimal-pair {ABX} task: analysis of the classical {MFC}/{PLP} pipeline},
    year={2013},
	url = {https://www.isca-archive.org/interspeech_2013/schatz13_interspeech.html},
	doi = {10.21437/Interspeech.2013-441},
	shorttitle = {Evaluating speech features with the minimal-pair {ABX} task},
	eventtitle = {Interspeech 2013},
	pages = {1781--1785},
	booktitle = {Interspeech 2013},
	publisher = {{ISCA}},
	author = {Schatz, Thomas and Peddinti, Vijayaditya and Bach, Francis and Jansen, Aren and Hermansky, Hynek and Dupoux, Emmanuel},
	urldate = {2024-10-02},
	date = {2013-08-25},
	langid = {english},
}

@article{kriegeskorte_deep_2015,
  title = {Deep Neural Networks: A New Framework for Modeling Biological Vision and Brain Information Processing},
  volume = {1},
  ISSN = {2374-4650},
  url = {http://dx.doi.org/10.1146/annurev-vision-082114-035447},
  DOI = {10.1146/annurev-vision-082114-035447},
  number = {1},
  journal = {Annual Review of Vision Science},
  publisher = {Annual Reviews},
  author = {Kriegeskorte,  Nikolaus},
  year = {2015},
  month = nov,
  pages = {417–446}
}

@article{huth_natural_2016,
  title = {Natural speech reveals the semantic maps that tile human cerebral cortex},
  volume = {532},
  ISSN = {1476-4687},
  url = {http://dx.doi.org/10.1038/nature17637},
  DOI = {10.1038/nature17637},
  number = {7600},
  journal = {Nature},
  publisher = {Springer Science and Business Media LLC},
  author = {Huth,  Alexander G. and de Heer,  Wendy A. and Griffiths,  Thomas L. and Theunissen,  Frédéric E. and Gallant,  Jack L.},
  year = {2016},
  month = apr,
  pages = {453–458}
}

@inproceedings{de_heer_kloots_human-like_2024,
	title = {Human-like Linguistic Biases in Neural Speech Models: Phonetic Categorization and Phonotactic Constraints in Wav2Vec2.0},
    year = {2024},
	url = {https://www.isca-archive.org/interspeech_2024/deheerkloots24_interspeech.html},
	doi = {10.21437/Interspeech.2024-2490},
	shorttitle = {Human-like Linguistic Biases in Neural Speech Models},
	eventtitle = {Interspeech 2024},
	pages = {4593--4597},
	booktitle = {Interspeech 2024},
	publisher = {{ISCA}},
	author = {De Heer Kloots, Marianne and Zuidema, Willem},
	urldate = {2024-10-01},
	date = {2024-09-01},
	langid = {english},
}

@article{waldis_holmes_2024,
	title = {Holmes {\textbackslash}ensuremath{\textbackslash}recorder {A} {Benchmark} to {Assess} the {Linguistic} {Competence} of {Language} {Models}},
	volume = {12},
	url = {https://aclanthology.org/2024.tacl-1.88/},
	doi = {10.1162/tacl_a_00718},
	urldate = {2025-11-18},
	journal = {Transactions of the Association for Computational Linguistics},
	author = {Waldis, Andreas and Perlitz, Yotam and Choshen, Leshem and Hou, Yufang and Gurevych, Iryna},
	year = {2024},
	note = {Place: Cambridge, MA
Publisher: MIT Press},
	pages = {1616--1647},
}

@misc{nickel_poincare_2017,
	title = {Poincar{\textbackslash}'e Embeddings for Learning Hierarchical Representations},
	url = {https://arxiv.org/abs/1705.08039v2},
	publisher = {{arXiv}},
	author = {Nickel, Maximilian and Kiela, Douwe},
	urldate = {2024-10-02},
	date = {2017-05-22},
	year={2017},
	langid = {english},
}

@article{pasad_what_2024,
	title = {What Do Self-Supervised Speech Models Know About Words?},
    journal = {Transactions of the Association for Computational Linguistics},
	volume = {12},
	issn = {2307-387X},
	url = {https://direct.mit.edu/tacl/article/doi/10.1162/tacl_a_00656/120586/What-Do-Self-Supervised-Speech-Models-Know-About},
	doi = {10.1162/tacl_a_00656},
	pages = {372--391},
	journaltitle = {Transactions of the Association for Computational Linguistics},
	author = {Pasad, Ankita and Chien, Chung-Ming and Settle, Shane and Livescu, Karen},
	urldate = {2024-10-01},
	year = {2024},
	langid = {english},
}

@article{dupoux_cognitive_2018,
title = {Cognitive science in the era of artificial intelligence: A roadmap for reverse-engineering the infant language-learner},
journal = {Cognition},
volume = {173},
pages = {43-59},
year = {2018},
issn = {0010-0277},
doi = {https://doi.org/10.1016/j.cognition.2017.11.008},
url = {https://www.sciencedirect.com/science/article/pii/S0010027717303013},
author = {Emmanuel Dupoux}
}

@inproceedings{chen2024sudden,
title={Sudden Drops in the Loss: Syntax Acquisition, Phase Transitions, and Simplicity Bias in {MLM}s},
author={Angelica Chen and Ravid Shwartz-Ziv and Kyunghyun Cho and Matthew L Leavitt and Naomi Saphra},
booktitle={The Twelfth International Conference on Learning Representations},
year={2024},
url={https://openreview.net/forum?id=MO5PiKHELW}
}

@article{Lavechin2025,
  title = {Simulating Early Phonetic and Word Learning Without Linguistic Categories},
  volume = {28},
  ISSN = {1467-7687},
  url = {http://dx.doi.org/10.1111/desc.13606},
  DOI = {10.1111/desc.13606},
  number = {2},
  journal = {Developmental Science},
  publisher = {Wiley},
  author = {Lavechin,  Marvin and de Seyssel,  Maureen and Titeux,  Hadrien and Wisniewski,  Guillaume and Bredin,  Hervé and Cristia,  Alejandrina and Dupoux,  Emmanuel},
  year = {2025},
  month = jan 
}

@article{frank2011computational,
  title={Computational models of early language acquisition},
  author={Frank, Michael C},
  journal={Current Opinion in Neurobiology},
  volume={21},
  number={3},
  pages={381--386},
  year={2011}
}

@article{kuhl2010brain,
  title={Brain mechanisms in early language acquisition},
  author={Kuhl, Patricia K},
  journal={Neuron},
  volume={67},
  number={5},
  pages={713--727},
  year={2010},
  publisher={Elsevier}
}

@article{martin2022plasticity,
  title={Plasticity of the language system in children and adults},
  author={Martin, Kelly C and Ketchabaw, W Tyler and Turkeltaub, Peter E},
  journal={Handbook of clinical neurology},
  volume={184},
  pages={397--414},
  year={2022},
  publisher={Elsevier}
}

@inproceedings{
park2025the,
title={The Geometry of Categorical and Hierarchical Concepts in Large Language Models},
author={Kiho Park and Yo Joong Choe and Yibo Jiang and Victor Veitch},
booktitle={The Thirteenth International Conference on Learning Representations},
year={2025},
url={https://openreview.net/forum?id=bVTM2QKYuA}
}

@article{poli2024modeling,
  title={Modeling the initial state of early phonetic learning in infants},
  author={Poli, Maxime and Schatz, Thomas and Dupoux, Emmanuel and Lavechin, Marvin},
  journal={Language Development Research},
  volume={5},
  number={1},
  year={2024}
}

@article{nakagi2025triple,
  title={Triple Phase Transitions: Understanding the Learning Dynamics of Large Language Models from a Neuroscience Perspective},
  author={Nakagi, Yuko and Tada, Keigo and Yoshino, Sota and Nishimoto, Shinji and Takagi, Yu},
  journal={arXiv preprint arXiv:2502.20779},
  year={2025}
}

@article{korchinski2025emergence,
  title={On the Emergence of Linear Analogies in Word Embeddings},
  author={Korchinski, Daniel J and Karkada, Dhruva and Bahri, Yasaman and Wyart, Matthieu},
  journal={arXiv preprint arXiv:2505.18651},
  year={2025}
}

@article{higgins2018towards,
  title={Towards a definition of disentangled representations},
  author={Higgins, Irina and Amos, David and Pfau, David and Racaniere, Sebastien and Matthey, Loic and Rezende, Danilo and Lerchner, Alexander},
  journal={arXiv preprint arXiv:1812.02230},
  year={2018}
}

@article{hu2024findings,
  title={Findings of the second BabyLM challenge: Sample-efficient pretraining on developmentally plausible corpora},
  author={Hu, Michael Y and Mueller, Aaron and Ross, Candace and Williams, Adina and Linzen, Tal and Zhuang, Chengxu and Cotterell, Ryan and Choshen, Leshem and Warstadt, Alex and Wilcox, Ethan Gotlieb},
  journal={arXiv preprint arXiv:2412.05149},
  year={2024}
}

@article{lavechin_can_2022,
	title = {Can statistical learning bootstrap early language acquisition? A modeling investigation},
	url = {https://osf.io/preprints/psyarxiv/rx94d/},
	shorttitle = {Can statistical learning bootstrap early language acquisition?},
    journal={PsyArXiv preprint psyarxiv/rx94d/},
	author = {Lavechin, Marvin and De Seyssel, Maureen and Titeux, Hadrien and Bredin, Hervé and Wisniewski, Guillaume and Cristia, Alejandrina and Dupoux, Emmanuel},
	urldate = {2024-12-19},
	year = {2022},
	note = {Publisher: {PsyArXiv}},
}

@article{oota2023joint,
  title={Joint processing of linguistic properties in brains and language models},
  author={Oota, SubbaReddy and Gupta, Manish and Toneva, Mariya},
  journal={Advances in Neural Information Processing Systems},
  volume={36},
  pages={18001--18014},
  year={2023}
}

@article{goldstein2022shared,
  title={Shared computational principles for language processing in humans and deep language models},
  author={Goldstein, Ariel and Zada, Zaid and Buchnik, Eliav and Schain, Mariano and Price, Amy and Aubrey, Bobbi and Nastase, Samuel A and Feder, Amir and Emanuel, Dotan and Cohen, Alon and others},
  journal={Nature neuroscience},
  volume={25},
  number={3},
  pages={369--380},
  year={2022},
  publisher={Nature Publishing Group US New York}
}

@inproceedings{seyssel_probing_2022,
	title = {Probing phoneme, language and speaker information in unsupervised speech representations},
	url = {http://arxiv.org/abs/2203.16193},
	doi = {10.21437/Interspeech.2022-373},
	pages = {1402--1406},
	booktitle = {Interspeech 2022},
	author = {Seyssel, Maureen de and Lavechin, Marvin and Adi, Yossi and Dupoux, Emmanuel and Wisniewski, Guillaume},
    year={2022},
	urldate = {2024-12-19},
	date = {2022-09-18},
	eprinttype = {arxiv},
	eprint = {2203.16193 [eess]},
}

@article{kachergis2022toward,
  title={Toward a “standard model” of early language learning},
  author={Kachergis, George and Marchman, Virginia A and Frank, Michael C},
  journal={Current Directions in Psychological Science},
  volume={31},
  number={1},
  pages={20--27},
  year={2022},
  publisher={Sage Publications Sage CA: Los Angeles, CA}
}

@misc{nguyen_zero_2020,
	title = {The Zero Resource Speech Benchmark 2021: Metrics and baselines for unsupervised spoken language modeling},
    year={2020},
	url = {http://arxiv.org/abs/2011.11588},
	doi = {10.48550/arXiv.2011.11588},
	shorttitle = {The Zero Resource Speech Benchmark 2021},
	number = {{arXiv}:2011.11588},
	publisher = {{arXiv}},
	author = {Nguyen, Tu Anh and Seyssel, Maureen de and Rozé, Patricia and Rivière, Morgane and Kharitonov, Evgeny and Baevski, Alexei and Dunbar, Ewan and Dupoux, Emmanuel},
	urldate = {2024-12-19},
	date = {2020-12-01},
	eprinttype = {arxiv},
	eprint = {2011.11588 [cs]},
}

@book{Pinker1996,
url = {https://doi.org/10.4159/9780674042179},
title = {Language Learnability and Language Development},
author = {Steven Pinker},
publisher = {Harvard University Press},
address = {Cambridge, MA and London, England},
doi = {doi:10.4159/9780674042179},
isbn = {9780674042179},
year = {1996},
lastchecked = {2025-05-21}
}

@article{Eilers1979,
  title = {Linguistic Experience and Phonemic Perception in Infancy: A Crosslinguistic Study},
  volume = {50},
  ISSN = {0009-3920},
  url = {http://dx.doi.org/10.2307/1129035},
  DOI = {10.2307/1129035},
  number = {1},
  journal = {Child Development},
  publisher = {JSTOR},
  author = {Eilers,  Rebecca E. and Gavin,  William and Wilson,  Wesley R.},
  year = {1979},
  month = mar,
  pages = {14}
}

@article{Christophe2008,
  title = {Bootstrapping Lexical and Syntactic Acquisition},
  volume = {51},
  ISSN = {1756-6053},
  url = {http://dx.doi.org/10.1177/00238309080510010501},
  DOI = {10.1177/00238309080510010501},
  number = {1–2},
  journal = {Language and Speech},
  publisher = {SAGE Publications},
  author = {Christophe,  Anne and Millotte,  Séverine and Bernal,  Savita and Lidz,  Jeffrey},
  year = {2008},
  month = mar,
  pages = {61–75}
}

@misc{rousso_tradition_2024,
	title = {Tradition or {Innovation}: {A} {Comparison} of {Modern} {ASR} {Methods} for {Forced} {Alignment}},
	shorttitle = {Tradition or {Innovation}},
	url = {http://arxiv.org/abs/2406.19363},
	doi = {10.48550/arXiv.2406.19363},
	urldate = {2025-09-12},
	publisher = {arXiv},
	author = {Rousso, Rotem and Cohen, Eyal and Keshet, Joseph and Chodroff, Eleanor},
	month = jun,
	year = {2024},
	note = {arXiv:2406.19363 [eess]
version: 1},
}

@article{orhan_detection_2025,
	title = {The detection of algebraic auditory structures emerges with self-supervised learning},
	volume = {21},
	url = {https://doi.org/10.1371/journal.pcbi.1013271},
	doi = {10.1371/journal.pcbi.1013271},
	number = {9},
	journal = {PLOS Computational Biology},
	author = {Orhan, Pierre and Boubenec, Yves and King, Jean-Rémi},
	month = sep,
	year = {2025},
	note = {Publisher: Public Library of Science},
	pages = {1--23},
}

@article{diego_simon_polar_2024,
	title = {A polar coordinate system represents syntax in large language models},
	volume = {37},
	url = {https://proceedings.neurips.cc/paper_files/paper/2024/hash/be36e50757bf9cd280aa74f89a7d1c23-Abstract-Conference.html},
	urldate = {2025-09-10},
	journal = {Advances in Neural Information Processing Systems},
	author = {Diego Simon, Pablo J. and d'Ascoli, Stéphane and Chemla, Emmanuel and Lakretz, Yair and King, Jean-Rémi},
	year = {2024},
	pages = {105375--105396},
}

@article{kuhl_early_2004,
	title = {Early language acquisition: cracking the speech code},
	volume = {5},
	copyright = {2004 Springer Nature Limited},
	issn = {1471-0048},
	shorttitle = {Early language acquisition},
	url = {https://www.nature.com/articles/nrn1533},
	doi = {10.1038/nrn1533},
	language = {en},
	number = {11},
	urldate = {2025-09-10},
	journal = {Nature Reviews Neuroscience},
	author = {Kuhl, Patricia K.},
	month = nov,
	year = {2004},
	note = {Publisher: Nature Publishing Group},
	pages = {831--843},
}

@inproceedings{evanson_language_2023,
	address = {Toronto, Canada},
	title = {Language acquisition: do children and language models follow similar learning stages?},
	shorttitle = {Language acquisition},
	url = {https://aclanthology.org/2023.findings-acl.773/},
	doi = {10.18653/v1/2023.findings-acl.773},
	urldate = {2025-08-19},
	booktitle = {Findings of the {Association} for {Computational} {Linguistics}: {ACL} 2023},
	publisher = {Association for Computational Linguistics},
	author = {Evanson, Linnea and Lakretz, Yair and King, Jean Rémi},
	editor = {Rogers, Anna and Boyd-Graber, Jordan and Okazaki, Naoaki},
	month = jul,
	year = {2023},
	pages = {12205--12218}
}

@article{saxe_mathematical_2019,
	title = {A mathematical theory of semantic development in deep neural networks},
	volume = {116},
	url = {https://www.pnas.org/doi/10.1073/pnas.1820226116},
	doi = {10.1073/pnas.1820226116},
	number = {23},
	urldate = {2022-09-14},
	journal = {Proceedings of the National Academy of Sciences},
	author = {Saxe, Andrew M. and McClelland, James L. and Ganguli, Surya},
	month = jun,
	year = {2019},
	note = {Publisher: Proceedings of the National Academy of Sciences},
	pages = {11537--11546}
}

@article{Gleitman1990,
  title = {The Structural Sources of Verb Meanings},
  volume = {1},
  ISSN = {1532-7817},
  url = {http://dx.doi.org/10.1207/s15327817la0101_2},
  DOI = {10.1207/s15327817la0101_2},
  number = {1},
  journal = {Language Acquisition},
  publisher = {Informa UK Limited},
  author = {Gleitman,  Lila},
  year = {1990},
  month = jan,
  pages = {3–55}
}

@article{Kohl1993,
  title = {Early linguistic experience and phonetic perception: implications for theories of developmental speech perception},
  volume = {21},
  ISSN = {0095-4470},
  url = {http://dx.doi.org/10.1016/S0095-4470(19)31326-9},
  DOI = {10.1016/s0095-4470(19)31326-9},
  number = {1–2},
  journal = {Journal of Phonetics},
  publisher = {Elsevier BV},
  author = {Kohl,  Patricia K.},
  year = {1993},
  month = jan,
  pages = {125–139}
}

@article{Jusczyk1993,
  title = {Infants´ Sensitivity to the Sound Patterns of Native Language Words},
  volume = {32},
  ISSN = {0749-596X},
  url = {http://dx.doi.org/10.1006/jmla.1993.1022},
  DOI = {10.1006/jmla.1993.1022},
  number = {3},
  journal = {Journal of Memory and Language},
  publisher = {Elsevier BV},
  author = {Jusczyk,  P.W. and Friederici,  A.D. and Wessels,  J.M.I. and Svenkerud,  V.Y. and Jusczyk,  A.M.},
  year = {1993},
  month = jun,
  pages = {402–420}
}

@article{Caucheteux2022,
  title = {Brains and algorithms partially converge in natural language processing},
  volume = {5},
  ISSN = {2399-3642},
  url = {http://dx.doi.org/10.1038/s42003-022-03036-1},
  DOI = {10.1038/s42003-022-03036-1},
  number = {1},
  journal = {Communications Biology},
  publisher = {Springer Science and Business Media LLC},
  author = {Caucheteux,  Charlotte and King,  Jean-Rémi},
  year = {2022},
  month = feb 
}

@inproceedings{Jawahar2019,
    title = "What Does {BERT} Learn about the Structure of Language?",
    author = "Jawahar, Ganesh  and
      Sagot, Beno{\^i}t  and
      Seddah, Djam{\'e}",
    editor = "Korhonen, Anna  and
      Traum, David  and
      M{\`a}rquez, Llu{\'i}s",
    booktitle = "Proceedings of the 57th Annual Meeting of the Association for Computational Linguistics",
    month = jul,
    year = "2019",
    address = "Florence, Italy",
    publisher = "Association for Computational Linguistics",
    url = "https://aclanthology.org/P19-1356/",
    doi = "10.18653/v1/P19-1356",
    pages = "3651--3657"
}

@book{Tesniere1959,
  address = {Paris},
  author = {Tesni{\`e}re, Lucien},
  publisher = {Klincksieck},
  title = {El{\'e}ments de Syntaxe Structurale},
  year = 1959
}

@inproceedings{Nivre2016,
    title = "{U}niversal {D}ependencies v1: A Multilingual Treebank Collection",
    author = "Nivre, Joakim  and
      de Marneffe, Marie-Catherine  and
      Ginter, Filip  and
      Goldberg, Yoav  and
      Haji{\v{c}}, Jan  and
      Manning, Christopher D.  and
      McDonald, Ryan  and
      Petrov, Slav  and
      Pyysalo, Sampo  and
      Silveira, Natalia  and
      Tsarfaty, Reut  and
      Zeman, Daniel",
    editor = "Calzolari, Nicoletta  and
      Choukri, Khalid  and
      Declerck, Thierry  and
      Goggi, Sara  and
      Grobelnik, Marko  and
      Maegaard, Bente  and
      Mariani, Joseph  and
      Mazo, Helene  and
      Moreno, Asuncion  and
      Odijk, Jan  and
      Piperidis, Stelios",
    booktitle = "Proceedings of the Tenth International Conference on Language Resources and Evaluation ({LREC}'16)",
    month = may,
    year = "2016",
    address = "Portoro{\v{z}}, Slovenia",
    publisher = "European Language Resources Association (ELRA)",
    url = "https://aclanthology.org/L16-1262/",
    pages = "1659--1666"
}

@inproceedings{Millet2022,
 author = {Millet, Juliette and Caucheteux, Charlotte and orhan, pierre and Boubenec, Yves and Gramfort, Alexandre and Dunbar, Ewan and Pallier, Christophe and King, Jean-Remi},
 booktitle = {Advances in Neural Information Processing Systems},
 editor = {S. Koyejo and S. Mohamed and A. Agarwal and D. Belgrave and K. Cho and A. Oh},
 pages = {33428--33443},
 publisher = {Curran Associates, Inc.},
 title = {Toward a realistic model of speech processing in the brain with self-supervised learning},
 url = {https://proceedings.neurips.cc/paper_files/paper/2022/file/d81ecfc8fb18e833a3fa0a35d92532b8-Paper-Conference.pdf},
 volume = {35},
 year = {2022}
}

@inproceedings{Silveira2014,
  year = {2014},
  author = {Natalia Silveira and Timothy Dozat and Marie-Catherine de
	  Marneffe and Samuel Bowman and Miriam Connor and John Bauer and
	  Christopher D. Manning},
  title = {A Gold Standard Dependency Corpus for {E}nglish},
  booktitle = {Proceedings of the Ninth International Conference on Language
    Resources and Evaluation (LREC-2014)}
}

@inproceedings{MullerEberstein2022,
    title = "Probing for Labeled Dependency Trees",
    author = {M{\"u}ller-Eberstein, Max  and
      van der Goot, Rob  and
      Plank, Barbara},
    editor = "Muresan, Smaranda  and
      Nakov, Preslav  and
      Villavicencio, Aline",
    booktitle = "Proceedings of the 60th Annual Meeting of the Association for Computational Linguistics (Volume 1: Long Papers)",
    month = may,
    year = "2022",
    address = "Dublin, Ireland",
    publisher = "Association for Computational Linguistics",
    url = "https://aclanthology.org/2022.acl-long.532/",
    doi = "10.18653/v1/2022.acl-long.532",
    pages = "7711--7726"
}

@inproceedings{Diego-Simon2024,
 author = {Diego-Sim\'{o}n, Pablo and D\textquotesingle Ascoli, St\'{e}phane and Chemla, Emmanuel and Lakretz, Yair and King, Jean-R\'{e}mi},
 booktitle = {Advances in Neural Information Processing Systems},
 editor = {A. Globerson and L. Mackey and D. Belgrave and A. Fan and U. Paquet and J. Tomczak and C. Zhang},
 pages = {105375--105396},
 publisher = {Curran Associates, Inc.},
 title = {A Polar coordinate system represents syntax in large language models},
 url = {https://proceedings.neurips.cc/paper_files/paper/2024/file/be36e50757bf9cd280aa74f89a7d1c23-Paper-Conference.pdf},
 volume = {37},
 year = {2024}
}

@inproceedings{Limisiewicz2021,
    title = "Introducing Orthogonal Constraint in Structural Probes",
    author = "Limisiewicz, Tomasz  and
      Mare{\v{c}}ek, David",
    editor = "Zong, Chengqing  and
      Xia, Fei  and
      Li, Wenjie  and
      Navigli, Roberto",
    booktitle = "Proceedings of the 59th Annual Meeting of the Association for Computational Linguistics and the 11th International Joint Conference on Natural Language Processing (Volume 1: Long Papers)",
    month = aug,
    year = "2021",
    address = "Online",
    publisher = "Association for Computational Linguistics",
    url = "https://aclanthology.org/2021.acl-long.36/",
    doi = "10.18653/v1/2021.acl-long.36",
    pages = "428--442"
}

@article{Yuan2019,
  title = {Bidirectional Transition-Based Dependency Parsing},
  volume = {33},
  ISSN = {2159-5399},
  url = {http://dx.doi.org/10.1609/aaai.v33i01.33017434},
  DOI = {10.1609/aaai.v33i01.33017434},
  number = {01},
  journal = {Proceedings of the AAAI Conference on Artificial Intelligence},
  publisher = {Association for the Advancement of Artificial Intelligence (AAAI)},
  author = {Yuan,  Yunzhe and Jiang,  Yong and Tu,  Kewei},
  year = {2019},
  month = jul,
  pages = {7434–7441}
}

@inproceedings{Hershcovich2018,
    title = "{U}niversal {D}ependency Parsing with a General Transition-Based {DAG} Parser",
    author = "Hershcovich, Daniel  and
      Abend, Omri  and
      Rappoport, Ari",
    editor = "Zeman, Daniel  and
      Haji{\v{c}}, Jan",
    booktitle = "Proceedings of the {C}o{NLL} 2018 Shared Task: Multilingual Parsing from Raw Text to Universal Dependencies",
    month = oct,
    year = "2018",
    address = "Brussels, Belgium",
    publisher = "Association for Computational Linguistics",
    url = "https://aclanthology.org/K18-2010/",
    doi = "10.18653/v1/K18-2010",
    pages = "103--112"
}

@inproceedings{Grunewald2021,
    title = "Applying Occam{'}s Razor to Transformer-Based Dependency Parsing: What Works, What Doesn{'}t, and What is Really Necessary",
    author = {Gr{\"u}newald, Stefan  and
      Friedrich, Annemarie  and
      Kuhn, Jonas},
    editor = "Oepen, Stephan  and
      Sagae, Kenji  and
      Tsarfaty, Reut  and
      Bouma, Gosse  and
      Seddah, Djam{\'e}  and
      Zeman, Daniel",
    booktitle = "Proceedings of the 17th International Conference on Parsing Technologies and the IWPT 2021 Shared Task on Parsing into Enhanced Universal Dependencies (IWPT 2021)",
    month = aug,
    year = "2021",
    address = "Online",
    publisher = "Association for Computational Linguistics",
    url = "https://aclanthology.org/2021.iwpt-1.13/",
    doi = "10.18653/v1/2021.iwpt-1.13",
    pages = "131--144"
}

@inproceedings{Tenney2019,
    title = "{BERT} Rediscovers the Classical {NLP} Pipeline",
    author = "Tenney, Ian  and
      Das, Dipanjan  and
      Pavlick, Ellie",
    editor = "Korhonen, Anna  and
      Traum, David  and
      M{\`a}rquez, Llu{\'i}s",
    booktitle = "Proceedings of the 57th Annual Meeting of the Association for Computational Linguistics",
    month = jul,
    year = "2019",
    address = "Florence, Italy",
    publisher = "Association for Computational Linguistics",
    url = "https://aclanthology.org/P19-1452/",
    doi = "10.18653/v1/P19-1452",
    pages = "4593--4601"
}

@INPROCEEDINGS{Pasad2021,
  author={Pasad, Ankita and Chou, Ju-Chieh and Livescu, Karen},
  booktitle={2021 IEEE Automatic Speech Recognition and Understanding Workshop (ASRU)}, 
  title={Layer-Wise Analysis of a Self-Supervised Speech Representation Model}, 
  year={2021},
  volume={},
  number={},
  pages={914-921},
  doi={10.1109/ASRU51503.2021.9688093}}

@article{Pasad2024,
    author = {Pasad, Ankita and Chien, Chung-Ming and Settle, Shane and Livescu, Karen},
    title = {What Do Self-Supervised Speech Models Know About Words?},
    journal = {Transactions of the Association for Computational Linguistics},
    volume = {12},
    pages = {372-391},
    year = {2024},
    month = {04},
    issn = {2307-387X},
    doi = {10.1162/tacl_a_00656},
    url = {https://doi.org/10.1162/tacl\_a\_00656},
    eprint = {https://direct.mit.edu/tacl/article-pdf/doi/10.1162/tacl\_a\_00656/2362252/tacl\_a\_00656.pdf},
}

@article{Marneffe2021,
    title = "{U}niversal {D}ependencies",
    author = "de Marneffe, Marie-Catherine  and
      Manning, Christopher D.  and
      Nivre, Joakim  and
      Zeman, Daniel",
    journal = "Computational Linguistics",
    volume = "47",
    number = "2",
    month = jun,
    year = "2021",
    address = "Cambridge, MA",
    publisher = "MIT Press",
    url = "https://aclanthology.org/2021.cl-2.11/",
    doi = "10.1162/coli_a_00402",
    pages = "255--308"
}

@inproceedings{Mcauliffe2017,
  title     = {Montreal Forced Aligner: Trainable Text-Speech Alignment Using Kaldi},
  author    = {Michael McAuliffe and Michaela Socolof and Sarah Mihuc and Michael Wagner and Morgan Sonderegger},
  year      = {2017},
  booktitle = {Interspeech 2017},
  pages     = {498--502},
  doi       = {10.21437/Interspeech.2017-1386},
  issn      = {2958-1796},
}

@article{Barrault2025,
  title = {Joint speech and text machine translation for up to 100 languages},
  volume = {637},
  ISSN = {1476-4687},
  url = {http://dx.doi.org/10.1038/s41586-024-08359-z},
  DOI = {10.1038/s41586-024-08359-z},
  number = {8046},
  journal = {Nature},
  publisher = {Springer Science and Business Media LLC},
  author = {Barrault,  Loïc and Chung,  Yu-An and Meglioli,  Mariano Coria and Dale,  David and Dong,  Ning and Duquenne,  Paul-Ambroise and Elsahar,  Hady and Gong,  Hongyu and Heffernan,  Kevin and Hoffman,  John and Klaiber,  Christopher and Li,  Pengwei and Licht,  Daniel and Maillard,  Jean and Rakotoarison,  Alice and Sadagopan,  Kaushik Ram and Wenzek,  Guillaume and Ye,  Ethan and Akula,  Bapi and Chen,  Peng-Jen and El Hachem,  Naji and Ellis,  Brian and Gonzalez,  Gabriel Mejia and Haaheim,  Justin and Hansanti,  Prangthip and Howes,  Russ and Huang,  Bernie and Hwang,  Min-Jae and Inaguma,  Hirofumi and Jain,  Somya and Kalbassi,  Elahe and Kallet,  Amanda and Kulikov,  Ilia and Lam,  Janice and Li,  Daniel and Ma,  Xutai and Mavlyutov,  Ruslan and Peloquin,  Benjamin and Ramadan,  Mohamed and Ramakrishnan,  Abinesh and Sun,  Anna and Tran,  Kevin and Tran,  Tuan and Tufanov,  Igor and Vogeti,  Vish and Wood,  Carleigh and Yang,  Yilin and Yu,  Bokai and Andrews,  Pierre and Balioglu,  Can and Costa-jussà,  Marta R. and undefinedelebi,  Onur and Elbayad,  Maha and Gao,  Cynthia and Guzmán,  Francisco and Kao,  Justine and Lee,  Ann and Mourachko,  Alexandre and Pino,  Juan and Popuri,  Sravya and Ropers,  Christophe and Saleem,  Safiyyah and Schwenk,  Holger and Tomasello,  Paden and Wang,  Changhan and Wang,  Jeff and Wang,  Skyler},
  year = {2025},
  month = jan,
  pages = {587–593}
}

@inproceedings{Biderman2023,
author = {Biderman, Stella and Schoelkopf, Hailey and Anthony, Quentin and Bradley, Herbie and O'Brien, Kyle and Hallahan, Eric and Khan, Mohammad Aflah and Purohit, Shivanshu and Prashanth, USVSN Sai and Raff, Edward and Skowron, Aviya and Sutawika, Lintang and Van Der Wal, Oskar},
title = {Pythia: a suite for analyzing large language models across training and scaling},
year = {2023},
publisher = {JMLR.org},
booktitle = {Proceedings of the 40th International Conference on Machine Learning},
articleno = {102},
numpages = {34},
location = {Honolulu, Hawaii, USA},
series = {ICML'23}
}

@article{Gao2020,
  title={The pile: An 800gb dataset of diverse text for language modeling},
  author={Gao, Leo and Biderman, Stella and Black, Sid and Golding, Laurence and Hoppe, Travis and Foster, Charles and Phang, Jason and He, Horace and Thite, Anish and Nabeshima, Noa and others},
  journal={arXiv preprint arXiv:2101.00027},
  year={2020}
}

@inproceedings{Baevski2020,
 author = {Baevski, Alexei and Zhou, Yuhao and Mohamed, Abdelrahman and Auli, Michael},
 booktitle = {Advances in Neural Information Processing Systems},
 editor = {H. Larochelle and M. Ranzato and R. Hadsell and M.F. Balcan and H. Lin},
 pages = {12449--12460},
 publisher = {Curran Associates, Inc.},
 title = {wav2vec 2.0: A Framework for Self-Supervised Learning of Speech Representations},
 url = {https://proceedings.neurips.cc/paper_files/paper/2020/file/92d1e1eb1cd6f9fba3227870bb6d7f07-Paper.pdf},
 volume = {33},
 year = {2020}
}

@article{Friedmann2015,
  title = {Critical period for first language: the crucial role of language input during the first year of life},
  volume = {35},
  ISSN = {0959-4388},
  url = {http://dx.doi.org/10.1016/j.conb.2015.06.003},
  DOI = {10.1016/j.conb.2015.06.003},
  journal = {Current Opinion in Neurobiology},
  publisher = {Elsevier BV},
  author = {Friedmann,  Naama and Rusou,  Dana},
  year = {2015},
  month = dec,
  pages = {27–34}
}

@article{Gomez1999,
  title = {Artificial grammar learning by 1-year-olds leads to specific and abstract knowledge},
  volume = {70},
  ISSN = {0010-0277},
  url = {http://dx.doi.org/10.1016/S0010-0277(99)00003-7},
  DOI = {10.1016/s0010-0277(99)00003-7},
  number = {2},
  journal = {Cognition},
  publisher = {Elsevier BV},
  author = {Gomez,  Rebecca L and Gerken,  LouAnn},
  year = {1999},
  month = mar,
  pages = {109–135}
}

@article{Boudreault2006,
  title = {Grammatical processing in American Sign Language: Age of first-language acquisition effects in relation to syntactic structure},
  volume = {21},
  ISSN = {1464-0732},
  url = {http://dx.doi.org/10.1080/01690960500139363},
  DOI = {10.1080/01690960500139363},
  number = {5},
  journal = {Language and Cognitive Processes},
  publisher = {Informa UK Limited},
  author = {Boudreault,  Patrick and Mayberry,  Rachel I.},
  year = {2006},
  month = aug,
  pages = {608–635}
}

@article{Parcollet2024,
title = {LeBenchmark 2.0: A standardized, replicable and enhanced framework for self-supervised representations of French speech},
journal = {Computer Speech \& Language},
volume = {86},
pages = {101622},
year = {2024},
issn = {0885-2308},
doi = {https://doi.org/10.1016/j.csl.2024.101622},
url = {https://www.sciencedirect.com/science/article/pii/S0885230824000056},
author = {Titouan Parcollet and Ha Nguyen and Solène Evain and Marcely {Zanon Boito} and Adrien Pupier and Salima Mdhaffar and Hang Le and Sina Alisamir and Natalia Tomashenko and Marco Dinarelli and Shucong Zhang and Alexandre Allauzen and Maximin Coavoux and Yannick Estève and Mickael Rouvier and Jerôme Goulian and Benjamin Lecouteux and François Portet and Solange Rossato and Fabien Ringeval and Didier Schwab and Laurent Besacier}
}

@inproceedings{Sashank2018,
title={On the Convergence of Adam and Beyond},
author={Sashank J. Reddi and Satyen Kale and Sanjiv Kumar},
booktitle={International Conference on Learning Representations},
year={2018},
url={https://openreview.net/forum?id=ryQu7f-RZ},
}

@article{Kingma2014,
  title={Adam: A method for stochastic optimization},
  author={Kingma, Diederik P and Ba, Jimmy},
  journal={arXiv preprint arXiv:1412.6980},
  year={2014}
}

@inproceedings{mortensen-etal-2016-panphon,
    title = "{P}an{P}hon: A Resource for Mapping {IPA} Segments to Articulatory Feature Vectors",
    author = "Mortensen, David R.  and
      Littell, Patrick  and
      Bharadwaj, Akash  and
      Goyal, Kartik  and
      Dyer, Chris  and
      Levin, Lori",
    editor = "Matsumoto, Yuji  and
      Prasad, Rashmi",
    booktitle = "Proceedings of {COLING} 2016, the 26th International Conference on Computational Linguistics: Technical Papers",
    month = dec,
    year = "2016",
    address = "Osaka, Japan",
    publisher = "The COLING 2016 Organizing Committee",
    url = "https://aclanthology.org/C16-1328/",
    pages = "3475--3484"
}

@article{gilkerson_mapping_2017,
	title = {Mapping the {Early} {Language} {Environment} {Using} {All}-{Day} {Recordings} and {Automated} {Analysis}},
	volume = {26},
	issn = {1058-0360, 1558-9110},
	url = {http://pubs.asha.org/doi/10.1044/2016_AJSLP-15-0169},
	doi = {10.1044/2016_AJSLP-15-0169},
	language = {en},
	number = {2},
	urldate = {2024-10-01},
	journal = {American Journal of Speech-Language Pathology},
	author = {Gilkerson, Jill and Richards, Jeffrey A. and Warren, Steven F. and Montgomery, Judith K. and Greenwood, Charles R. and Kimbrough Oller, D. and Hansen, John H. L. and Paul, Terrance D.},
	month = may,
	year = {2017},
	pages = {248--265},
}

@article{landauer1997solution,
  title={A solution to Plato's problem: The latent semantic analysis theory of acquisition, induction, and representation of knowledge.},
  author={Landauer, Thomas K and Dumais, Susan T},
  journal={Psychological review},
  volume={104},
  number={2},
  pages={211},
  year={1997},
  publisher={American Psychological Association}
}

@article{frank2023large,
  title = {Large language models as models of human cognition},
  url = {http://dx.doi.org/10.31234/osf.io/wxt69},
  DOI = {10.31234/osf.io/wxt69},
  publisher = {Center for Open Science},
  author = {Frank,  Michael C.},
  year = {2023},
  journal={PsyarXiv},
  month = jul 
}

@article{dubois2016exploring,
  title={Exploring the early organization and maturation of linguistic pathways in the human infant brain},
  author={Dubois, Jessica and Poupon, Cyril and Thirion, Bertrand and Simonnet, Hina and Kulikova, Sofya and Leroy, Fran{\c{c}}ois and Hertz-Pannier, Lucie and Dehaene-Lambertz, Ghislaine},
  journal={Cerebral Cortex},
  volume={26},
  number={5},
  pages={2283--2298},
  year={2016},
  publisher={Oxford University Press}
}

@unpublished{evanson2025emergence,
  author       = {Linnea Evanson and Christine Bulteau and Mathilde Chipaux and Georg Dorfmüller and Sarah Ferrand-Sorbets and Emmanuel Raffo and Sarah Rosenberg and Pierre Bourdillon and Jean-Rémi King},
  title        = {Emergence of Language in the Developing Brain},
  year         = {2025},
  month        = {May 12},
  note         = {Preprint, AI at Meta \& Fondation Adolphe de Rothschild Hospital. Available at: \url{https://ai.meta.com/research/publications/emergence-of-language-in-the-developing-brain/}},
  howpublished = {Preprint}
}

@article{vong2024grounded,
  title={Grounded language acquisition through the eyes and ears of a single child},
  author={Vong, Wai Keen and Wang, Wentao and Orhan, A Emin and Lake, Brenden M},
  journal={Science},
  volume={383},
  number={6682},
  pages={504--511},
  year={2024},
  publisher={American Association for the Advancement of Science}
}

@inproceedings{hewitt_structural_2019,
	address = {Minneapolis, Minnesota},
	title = {A {Structural} {Probe} for {Finding} {Syntax} in {Word} {Representations}},
	url = {http://aclweb.org/anthology/N19-1419},
	doi = {10.18653/v1/N19-1419},
	language = {en},
	urldate = {2024-10-01},
	booktitle = {Proceedings of the 2019 {Conference} of the {North}},
	publisher = {Association for Computational Linguistics},
	author = {Hewitt, John and Manning, Christopher D.},
	year = {2019},
	pages = {4129--4138},
}

@article{kemp_discovery_2008,
	title = {The discovery of structural form},
    year = {2008},
	volume = {105},
	url = {https://www.pnas.org/doi/10.1073/pnas.0802631105},
	doi = {10.1073/pnas.0802631105},
	number = {31},
	urldate = {2024-10-12},
	journal = {Proceedings of the National Academy of Sciences},
	author = {Kemp, Charles and Tenenbaum, Joshua B.},
	month = aug,
	note = {Publisher: Proceedings of the National Academy of Sciences},
	pages = {10687--10692},
}

@misc{jha_harnessing_2025,
	title = {Harnessing the {Universal} {Geometry} of {Embeddings}},
	url = {http://arxiv.org/abs/2505.12540},
	doi = {10.48550/arXiv.2505.12540},
	urldate = {2025-08-20},
	publisher = {arXiv},
	author = {Jha, Rishi and Zhang, Collin and Shmatikov, Vitaly and Morris, John X.},
	month = may,
	year = {2025},
	note = {arXiv:2505.12540 [cs]
version: 1},
}

@misc{huh_platonic_2024,
	title = {The {Platonic} {Representation} {Hypothesis}},
	url = {http://arxiv.org/abs/2405.07987},
	doi = {10.48550/arXiv.2405.07987},
	urldate = {2025-08-20},
	publisher = {arXiv},
	author = {Huh, Minyoung and Cheung, Brian and Wang, Tongzhou and Isola, Phillip},
	month = jul,
	year = {2024},
	note = {arXiv:2405.07987 [cs]},
}

@article{friedmann2021growing,
  title={Growing trees: The acquisition of the left periphery},
  author={Friedmann, Naama and Belletti, Adriana and Rizzi, Luigi},
  journal={Glossa: a journal of general linguistics},
  volume={6},
  number={1},
  year={2021},
  publisher={Open Library of Humanities}
}

@misc{ji_predicting_2022,
	title = {Predicting within and across language phoneme recognition performance of self-supervised learning speech pre-trained models},
	url = {http://arxiv.org/abs/2206.12489},
	doi = {10.48550/arXiv.2206.12489},
	urldate = {2025-09-23},
	publisher = {arXiv},
	author = {Ji, Hang and Patel, Tanvina and Scharenborg, Odette},
	month = jun,
	year = {2022},
	note = {arXiv:2206.12489 [eess]},
}

\appendix
\renewcommand{\figurename}{Supplementary Figure}
\renewcommand{\thefigure}{\Alph{figure}}
\setcounter{figure}{0}

\renewcommand{\tablename}{Table}
\renewcommand{\thetable}{\Alph{table}}
\setcounter{table}{0}

\newpage

\section{Appendix}
\begin{table}[h]
  \centering
  \begin{tabular}{{cccc}}
    \toprule
    Name & Modality & Annotation\\ 
    \midrule
    UD--EWT        & text  & UD (\textbf{\textsc{m}})\\
    WordNet--Nouns     & text & WordNet (\textbf{\textsc{m}})\\
    UD--EWT--TTS   & speech & UD (\textbf{\textsc{m}})\,+ Phonemes (\textbf{\textsc{a}})\\
    WordNet--Nouns--TTS    & speech & WordNet (\textbf{\textsc{m}})\\
    \bottomrule
  \end{tabular}
  \caption{Datasets used to train syntactic, lexical semantic and phonological probes. Annotations of linguistic structures can be manual or automatic, marked by \textsc{m} and \textsc{a} respectively.}
  \label{tab:data}
\end{table}

\begin{table}[ht]
    \centering
    \begin{tabular}{lcccccc}
        \toprule
        \textbf{Structure} & \textbf{Model} & \textbf{Learning rate} &\textbf{Dim} & \textbf{Epochs} & \textbf{Probe init} & \textbf{batch size} \\
        \midrule
        phonemic & Wav2vec 2.0 & 0.00001 & 768 &  100 & 0.00001 & 1 set of 200 phonemes  \\
        semantic & Wav2vec 2.0 30-94M  & 0.0000075 & 200 & 200 & 0.00001 & 300 set of 12 words \\
        semantic & Wav2vec 2.0 315M &  0.0005 & 200 & 200 & 0.00001 & 300 set of 12 words \\
        semantic & pythia & 0.0000025 & 200 & 300 & 0.00001 & 300 set of 12 words \\
        semantic & Llama 2.0 & 0.00005 & 200 & 300 & 0.00001 & 300 set of 12 words \\
        syntactic & Wav2vec 2.0 & 0.00001  & 200 & 2 & 0.00001 & 300 sentences  \\
        syntactic & pythia & 0.000025  & 500 & 1 & 0.00001 & 150 sentences  \\
        syntactic & Llama 2.0 & 0.00001  & 200 & 1 & 0.00001 &  300 sentences\\
        \bottomrule
    \end{tabular}
    \caption{Parameters of the linear probe. All probes are trained with the AMSGrad variant \citep{Sashank2018} of the Adam optimizer \citep{Kingma2014}
    For each combination of (model, task), we perform a grid search across learning rate, spanning linearly 20 values from 0.0000001 to 0.01, and report the best learning rate. 
    Unlike its sensitivity to learning rate, the probe is more robust to batch size and probe initialization value.
    The probe is randomly initialized from a uniform distribution between -0.00001 and 0.00001.
    For each model, all hyperparameters are constant across layers. When the model size is not given, it means that the same hyperparameters were used across all model sizes.}
    \label{tab:probing_hyperparameters}
\end{table}

\begin{table}[ht]
    \centering
    \begin{tabular}{lccccc}
        \toprule
         \textbf{Training Dataset} & \textbf{Model Size} & \textbf{Training Steps} & \textbf{Checkpoints} & \textbf{Hugging face repository}  \\
        \midrule
        Audioset & 30M  & 100k steps & \checkmark & \makecell[l]{NDEM/model-wav2vec2\_type-tiny\\ \_data-audiosetfilter\_version-0} \\
        Audioset & 94M  & 100k steps  & \checkmark & \makecell[l]{NDEM/model-wav2vec2\_type-base\\ \_data-audiosetfilter\_version-2} \\
        FMA & 30M  & 100k steps  & \checkmark & \makecell[l]{NDEM/model-wav2vec2\_type-tiny\\ \_data-fma\_version-0}\\
        FMA & 94M  & 100k steps  & \checkmark & \makecell[l]{NDEM/model-wav2vec2\_type-base\\ \_data-fma\_version-2}\\
        LeBench & 315M  & 200k steps & x & LeBenchmark/wav2vec2-FR-7K-large \\
        Librispeech & 30M  & 100k steps  & \checkmark & \makecell[l]{NDEM/model-wav2vec2\_type-tiny\\ \_data-librispeech\_version-0}  \\
        Librispeech & 94M  & 100k steps  & \checkmark & \makecell[l]{NDEM/model-wav2vec2\_type-base\\ \_data-librispeech\_version-4} \\
        Librispeech & 94M  & 400k steps  & \checkmark & \makecell[l]{NDEM/model-wav2vec2\_type-base\\ \_data-librispeech\_version-2} \\
        Librispeech & 94M  & 400k steps  & x  & facebook/wav2vec2-base\\
        Librispeech & 315M  & 200k steps & x & facebook/wav2vec2-large \\
        The Pile & \makecell[l]{70M-1.4B}  & 143k steps & \checkmark & EleutherAI/pythia-Xb \\
        Meta Internal corpus & 7B  & 500k steps & x & meta-llama/Llama-2-7b\\
        Meta Internal corpus & 13B  & 500k steps & x & meta-llama/Llama-2-13b\\

        \bottomrule
    \end{tabular}
    \caption{Set of audio and text models along with their training dataset, the model size, total number of training steps, and availability of checkpoints. Control audio models were taken from the following paper: \cite{orhan_detection_2025}.}
    \label{tab:model_comparison}
\end{table}

\begin{table}[ht]
    \centering
    \begin{tabular}{lcc}
    \toprule
    \textbf{Model} & \textbf{Training data} &\textbf{Semantic score (pearson correlation)} \\
    \midrule
    Wav2vec 2.0 30M & environmental sound & 0.044\\
    Wav2vec 2.0 94M & environmental sound & 0.065\\
    Wav2vec 2.0 30M & music & 0.080\\
    Wav2vec 2.0 94M & music & 0.070\\
    Wav2vec 2.0 315M & French & 0.126\\
    Wav2vec 2.0 30M & English & 0.08\\
    Wav2vec 2.0 94M & English & 0.104\\
    Wav2vec 2.0 315M & English & 0.147\\
    Pythia 70M & The Pile & 0.118 \\
    Pythia 160M & The Pile & 0.166 \\
    Pythia 410M & The Pile & 0.22 \\
    Pythia 1B & The Pile & 0.27 \\
    Pythia 1.4B & The Pile & 0.29 \\
    Llama 2 7B & Meta Internal Corpus & 0.42 \\
    Llama 2 13B & Meta Internal Corpus & 0.415 \\
    \end{tabular}
    \caption{Scores of the semantic probe, (max pearson scores across layers)}
    \label{tab:semantic}
\end{table}

\subsection*{Syntax Dataset}

Our generation pipeline will be made available in the form of the ALS (Auditory Linguistic Structure) python repository upon acceptance.
This pipeline takes as inputs an annotated text corpus (ConLLu format), and synthesizes the equivalent auditory dataset, along with a precise temporal alignment.
We make available a cluster-friendly version of this pipeline, which allows the reproduction of the dataset in less than one hour.

\paragraph{Speech synthesis}
For each sentence in the English-Web-Treebank (EWT) Universal Dependencies Dataset, we synthesize a voice clip using the ``seamless-m4t-v2-large'' deep learning model.
First, we remove from each sentence any remaining contractions using the contractions python package.
Removing contractions allows the Force Aligner to separate in time contracted words, (like You're) that would be otherwise recognized as a single event. 
Second, for each sentence, the text is used as input to a ``seamless-m4t-v2-large'' model, with default English speaker voice.
We save the generated sounds as a WAV file with a 16000 Hz sampling rate.

\paragraph{Alignment}
Forced Alignment (FA) is a necessary step to know when each word in a sentence occurs in its corresponding synthesized voice clip. 
FA was conducted using the Montreal Forced Aligner (MFA) model, employing the \texttt{english\_us\_arpa} corpus and dictionary.
To quantify the effect of different aligners, we also reproduced our analyses with the NeMo Forced Aligner. 
This aligner produced a worse alignment when manually checked, in accordance with \citet{rousso_tradition_2024} which compared MFA with other aligners and found MFA to outperform.
Indeed, everything else fixed, the syntactic and semantic scores to the NeMo-aligned corpus were poorer, which indicates measurement error induced by the alignment. 
Despite the great care we took for this forced alignment, all of our metrics remain a lower bound estimate of the true model's abilities, as there will necessarily be some alignment errors in the process.

The Montreal Forced Aligner does not map exactly each word in its input text to each annotated word in the input corpus.
Some composite words can indeed be split by MFA.
To avoid silent misalignment mistakes, we run a simple iterative merging algorithm to group set of split words.
This procedure leaves us with a beginning and an end time for every annotated word in the corpus.

Finally, we correct the information of the input corpus (ConLLu file), in the case of contractions.
Decontracted sets of words are kept split as several elements (this should be done already in the ConLLu), while for re-contracted word (as can't becoming cannot), we keep only the first ConLLu token of the set of tokens associated with the contractions.

\subsection*{Semantic Dataset}

\paragraph{Semantic category}
In the context of WordNet, we defined a semantic category as the set composed of a synset and all synsets that can reach this synset through hypernymy relationship (Fig \ref{fig:knnScores} in appendix).
For example, the noun graph root is ``entity.n.01'', and therefore all elements in the graph are part of the category ``entity.n.01''.
Similarly, all synsets below ``mammal.n.01'' are part of the ``mammal'' category, but ``animal.n.01'' which is a hypernym of ``mammal.n.01'' is not.

\paragraph{Dataset generation}
The mapping between WordNet synsets and the set of nouns is neither surjective nor injective.
Words are often polysemic, such that preserving words associated to a single synset would restrict our investigation to a small subgraph of the WordNet graph.
Conversely, synsets are associated with multiple sets of words (lemmas).
To deal with these two constraint, we can nevertheless take advantage of the fact that some synsets are over-precise and characterize a very particular use of the words. 
For example, time is associated to 15 synsets, among which: time.n.05: 'the continuum of experience in which events pass from the future through the present to the past', and time.n.02: 'an indefinite period (usually marked by specific attributes or activities)'.

Consequently, to generate a dataset of pairs (synset,lemma) from the WordNet graph, we applied the following heuristic. (function wordnet\_unisemic\_wordlist in the ALS package).
First, we collected all synsets in the WordNet noun graph.
Each synset is associated with a list of lemmas (each lemma is a word or multiple words).
We then repeated the following procedure:
Given this list of synset, we first, whenever it was possible, assigned to a synset all lemmas for which no other synset has this lemma in its list.
We then removed from the list of lemmas associated to this synset all lemmas that were in the list of other synsets.
This allowed to repeat the procedure, as some lemmas then became assigned to a unique synset.
In short, this first procedure selects for all synsets the lemma that is most unique to this synset.

This procedure still left a large number of polysemic lemmas, each associated with multiple synsets.
Among those synsets, there are leaf synsets which have no hyponyms and are therefore very precise sense.
We consequently pruned these leaf synsets out (removing a total of 9593 leafs synsets).
The rationale was that the non-contextual evaluation of the lexical semantics of a polysemic word should rather be on its general sense than its precise and context-specific sense.
Finally, we preserved in the graph the synset whose remaining list had at least one lemma associated with only one synset.
Given this subset of synsets and their associated list of unique lemmas, we ultimately selected the lemma that had the highest word frequency according to the wordfreq package.

\paragraph{Cross-validation folds for Wordnet}
Next, we divided this large graph into train and test sets, using a splitting strategy that ensured that for each semantic category (defined as a synset and all its hyponyms), $20\%$ of its members were in the test set.
To be able to quantify how well each semantic category was coded in the model, we had to put in the test folds a subset of the words for each category.
This assignment was not trivial, and we consequently came up with a simulated annealing procedure to obtain it.
This procedure minimized a loss function that measures for all categories, the percentage of its elements in the test set.
To be precise, the cost is defined as: $\sum_j\sum_{i\in C_j}(\delta_{i\in T}-0.2|C_j|)^2$ where $\delta_{i \in T}$ is 1 if the synset i is in the test set, and $C_j$ is the set of element in the category j, and the first sum iterates across all categories.
Running a simulated annealing procedure involves randomly changing an element to be part of the training or testing set.
This change is kept if the cost decreases ($\Delta cost<0$) or if we sample a random number smaller than $\exp(-(\Delta cost)/\tau)$. $\tau$ controls the temperature and is progressively cooled.
We run this simulated annealing procedure for 1000000 iterations with a cooling rate of 0.999995.

\paragraph{Restriction to the words appearing in librispeech}
In this work, we extensively compare audio and text models.
Consequently, we decided to restrict our evaluation to words that appeared at least once in the Librispeech dataset.
In the case of synsets associated with multiple words, we preserved only synsets for which all words appeared in the Librispeech dataset.
The final dataset is consequently taken from the output of the cross-validation procedure and filtered to preserve only nouns appearing in the Librispeech dataset.

This procedure provided a reasonably large sample of (synset,lemma) pairs: 35,352, while avoiding synsets with major polysemic issues.




\subsection*{Emergence of semantic category does not replicate findings in toy models}
The sequential emergence of increasingly complex semantic abilities is a property that was initially predicted in toy models \cite{saxe_exact_2014,saxe_mathematical_2019}. 
In essence, this toy model predicts that larger semantic classes should emerge before precise semantic classes.
We adapted these semantic classes by defining semantic categories in the Wordnet hierarchy (Fig \ref{fig:knnScores} in appendix).
To confirm whether these findings extended to transformer models, we then measured whether semantic categories emerged in the order of their size.
Surprisingly, we found that this was not the case: class size did not predict the time of emergence of the lexical class (Fig \ref{fig:knnScores} in appendix). 
For example, the separation of the main mammal subclasses, like ``primate'', ``marsupial'', ``rodent'', ``carnivore'', did not precede the separation of carnivore subclasses like ``dog'' or ``feline''.
On the other hand, this supplementary analysis showed that semantic classes indeed exist, and improve with model size (Fig \ref{fig:knnScores} in appendix), which strengthens our conclusions on the existence of a semantic graph in language models.

\subsection*{English exposed model instantiate partial semantic better than control models}
To demonstrate that an English exposed model instantiate partial semantics in a significantly better way than a control model, we measured how well they could classify each semantic category. 
Remarkably, across most (167, 89\%) large semantic categories, the English-exposed model had a better categorization score than control models. 
We show a scatter plot comparing these scores for  early (5) and late (10) layers (Fig~\ref{fig:sup_scatter_control_semantic} in the appendix panel A-B).
This difference increased through the layers of the model, peaking in layer 10 (Fig~\ref{fig:sup_scatter_control_semantic} in the appendix, panel C).
Here, we focus on large categories with more than 100 words to show that this is driven by major semantic categories (a total of 187 categories), but this was also true when we looked at all of the 35532 semantic categories in the dataset.

\subsection*{UMAP visualizations confirm evaluations of semantic scores}

We plot two UMAP visualisations before and after projection with the 200-D probe (Fig.~\ref{fig:sup_umap} in appendix). 
Before projection, the space is only roughly divided into large semantic categories, presenting many small clusters without a clear organisation. After projection, the space is well organised into continuous clusters mirroring the WordNet tree hierarchy.
Consequently, this unsupervised visualisation demonstrates that the WordNet tree is only instantiated in a subspace of the model activity, which is found by our probe. 

\subsection*{Additional evaluation metrics}

\begin{figure}[h]
  \centering
  \includegraphics[width=0.99\textwidth]{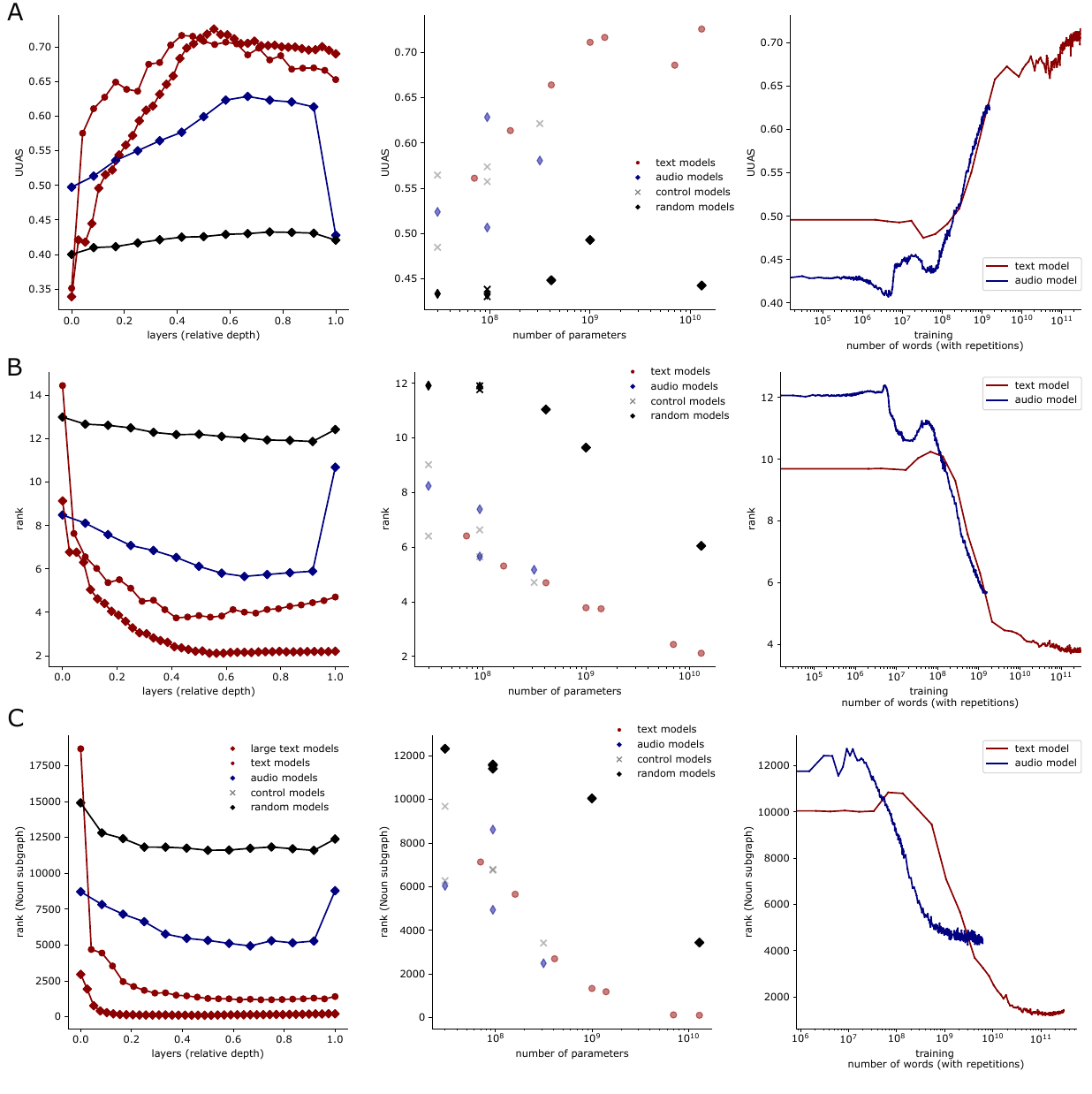}
\caption{A: Same as Fig.\ref{fig:syntactic}, but evaluating the Unlabeled Undirected Attachment Score (UUAS)
         B. Same as Fig.\ref{fig:syntactic}, but evaluating the rank score.
         C: Same as Fig.\ref{fig:semantic}, but evaluating the rank score}
\label{fig:sup_syntacticRankUUAS}
\end{figure}

We provide in Fig.~\ref{fig:sup_syntacticRankUUAS} additional results of the same syntactic and semantic probe but evaluated with two different metrics from the literature.
The Unlabeled Undirected Attachment Score (UAS) measures the percentage of correct syntactic edges recovered by the probe \citep{hewitt_structural_2019}.
The rank score divides for each node $i$ all other nodes into a positive ($P(i)$) set and negative ($N(i)$) set, depending on whether they are connected to $i$ or not in the graph (semantic or syntactic).
The rank score of this node is then the average rank of all positive nodes when sorting all nodes by increasing distance from $i$ \citep{nickel_poincare_2017}.
The final rank score of a probe is then the median of these ranks scores across all nodes \citep{nickel_poincare_2017}.

\begin{figure}[h]
  \centering
  \includegraphics[width=0.99\textwidth]{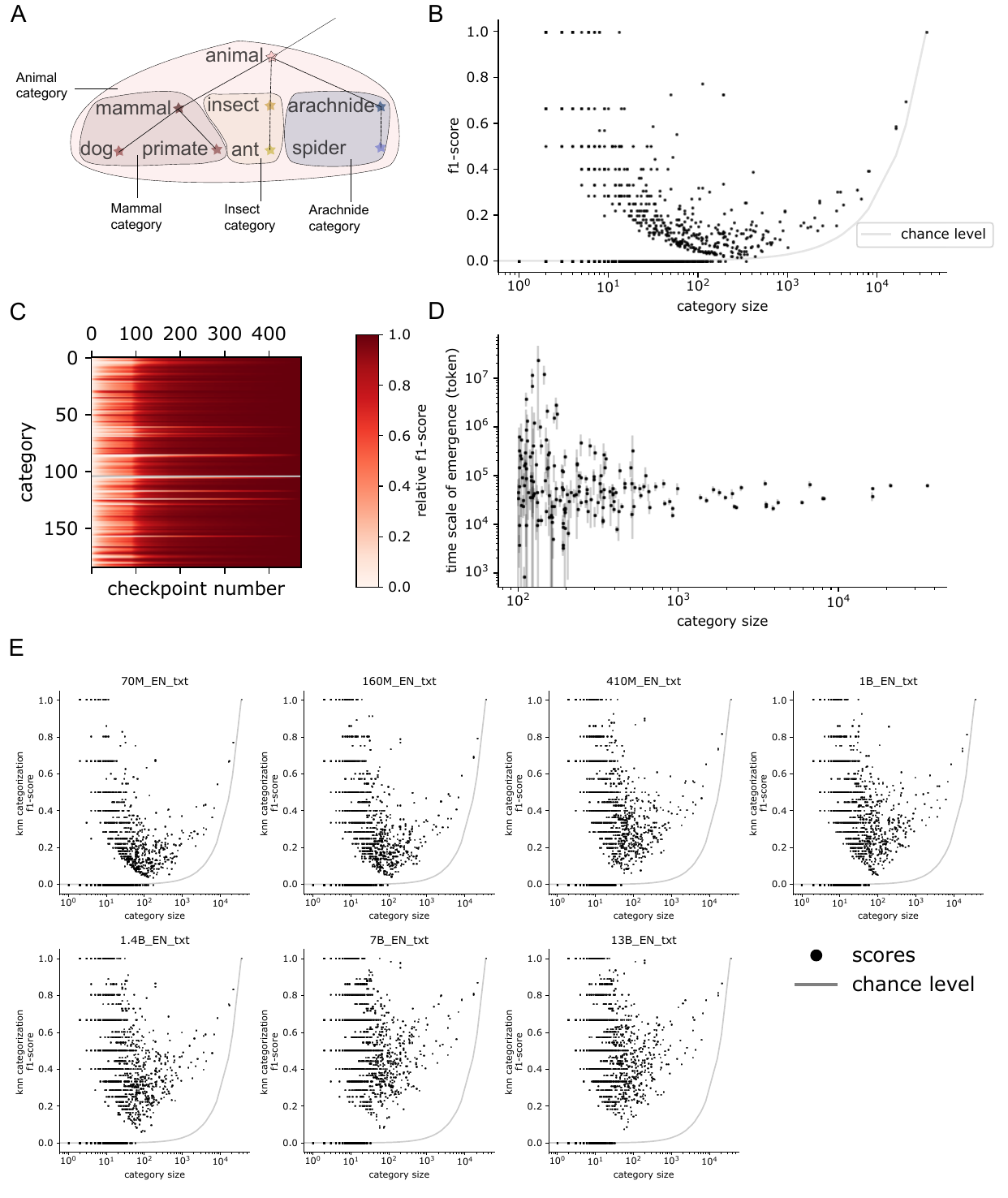}
\caption{A: Schematic of the definition of a category in a Wordnet hierarchy. B: f1-score of a 5-Nearest neighbor classification procedure, as a function of category size for a pretrained base Wav2vec 2.0 model.
C: f1-score as a function of the checkpoint.
D: Time scale of emergence as a function of the category size. Semantic categories do not form in the model in order of their size.
E: f1-score for all language models in this study (Pythia suite from 70M to 1.4 billion parameters and Llama 2.0 large language models).
}
\label{fig:knnScores}
\end{figure}

\begin{figure}[h]
  \centering
  \includegraphics[width=0.99\textwidth]{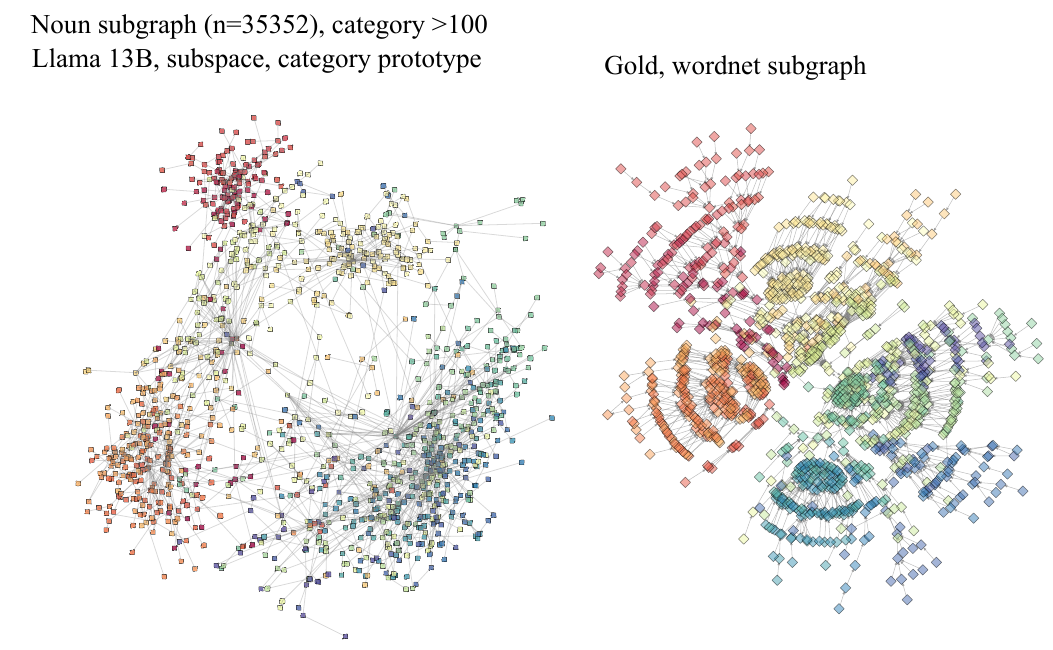}
\caption{Left: 2D projection of the noun subgraph probed from a Llama 2.0 models.
Right: gold WordNet noun subgraph.
}
\label{fig:sup_semanticWordnet}
\end{figure}

\begin{figure}[h]
  \centering
  \includegraphics[width=0.99\textwidth]{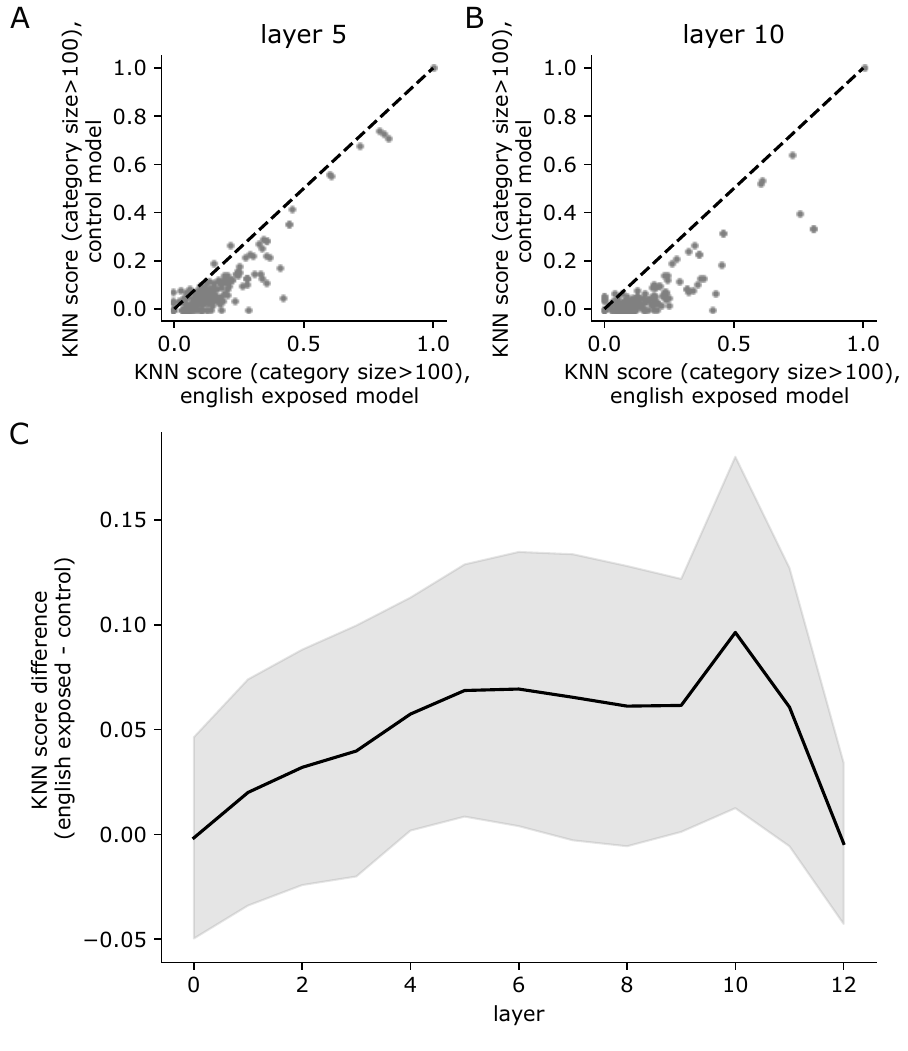}
\caption{A-B: Scatter plot of the f1-score of large semantic categories (more than 100 synsets) for an English-exposed Wav2vec 2.0 base model against a control model (Wav2vec 2.0 trained on environmental sounds) for layer 5 (A) and layer 10 (B).
        C: Difference of f1-score of the KNN classification between the English-exposed and control model, averaged across categories (black line) and with standard deviation (grey shadow) as a function of the layer.
}
\label{fig:sup_scatter_control_semantic}
\end{figure}

\begin{figure}[h]
  \centering
  \includegraphics[width=0.99\textwidth]{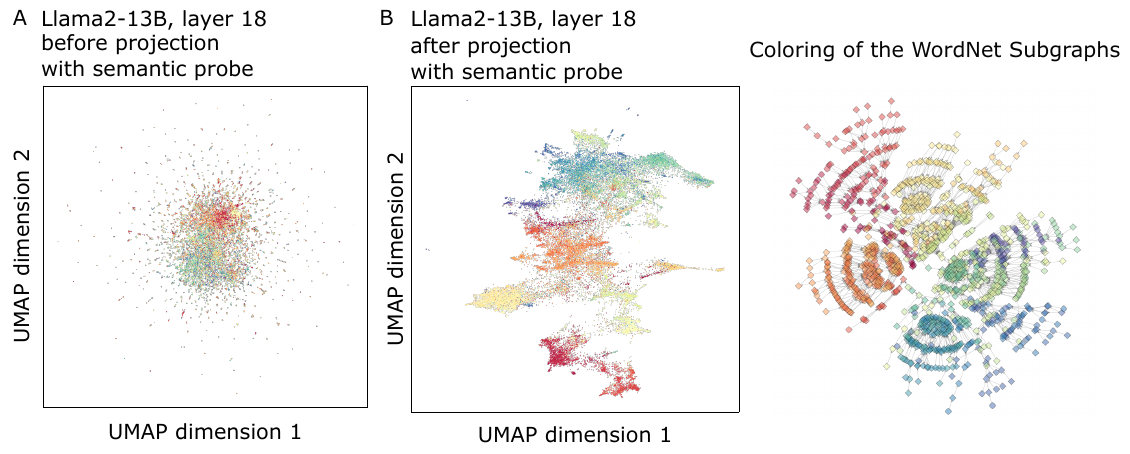}
\caption{A: UMAP visualization of the noun subgraph from the activity of a Llama 2.0 model.
B: UMAP visualization after projection of a Llama 2.0 model with a 200-D semantic probe.
}
\label{fig:sup_umap}
\end{figure}

\clearpage
\begin{figure}[h]
  \centering
  \includegraphics[width=0.99\textwidth]{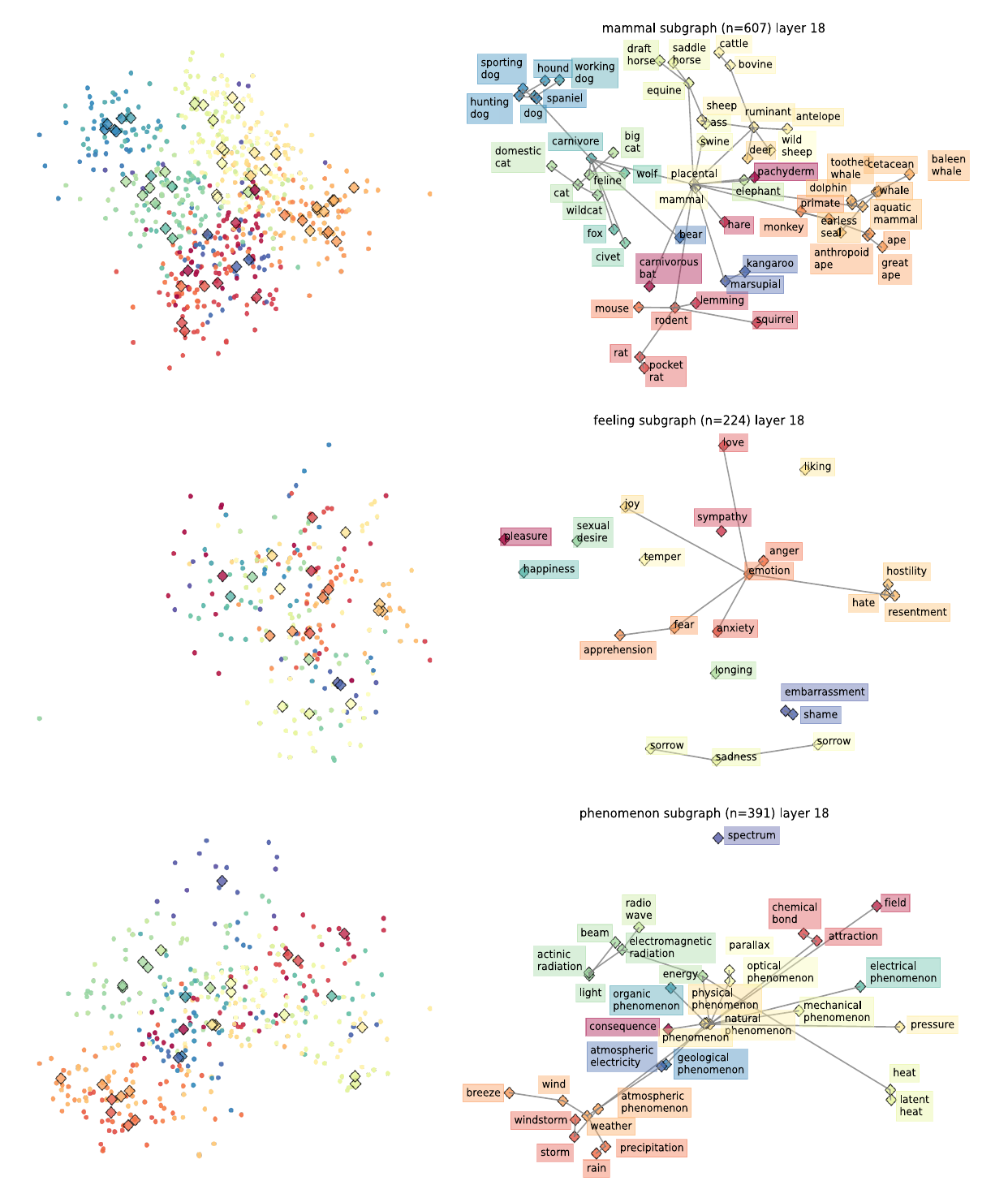}
\caption{Llama\_13B 2D probing of mammal, feeling and phenomenon subgraphs of Wordnet
}
\label{fig:sup_semantic1}
\end{figure}

\clearpage

\begin{figure}[h]
  \centering
  \includegraphics[width=0.99\textwidth]{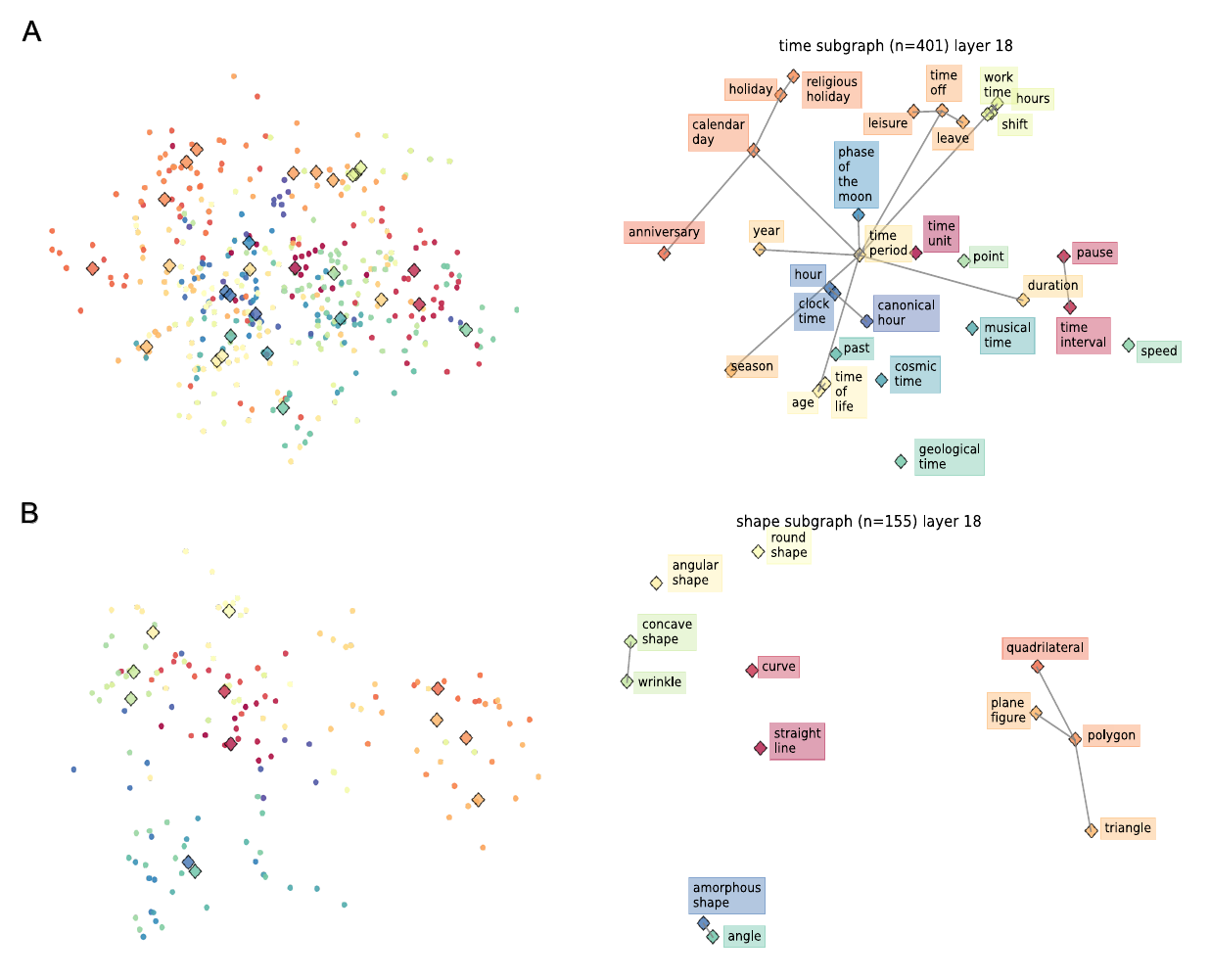}
\caption{Llama\_13B 2D probing of time and shape subgraphs of Wordnet
}
\label{fig:sup_semantic2}
\end{figure}

\subsection*{Single unit analyses of semantic and syntactic subspaces}

To gain some further insights on how single units interact in the coding of the semantic and syntactic subspace, we additionally ran 3 analyses on the syntactic and semantic probe of a Llama2 model.

We restricted these analyses to model layers for which the semantic and syntactic scores were above 80

First, to understand if two subspaces are aligned, we propose to define an alignment score as the mean square cosines of principal angles between two subspaces (Ake et al, 1973) (i,j):
Second, we studied the norm of each unit's probe weights.
Third, we quantified how well each unit's activity could be predicted by the semantic and syntactic subspace activity.

We first wanted to quantify whether semantic and syntactic subspaces recruited a similar pattern of units in the model. To do so, we computed the alignment between the left singular vector of each probe.
More precisely, we first compute the left singular vector of each probe (over the model units).
These vectors quantify how much each unit contributes to the activity of a particular subspace direction. Because they form an orthonormal basis from the same space, we can then compute an alignment score  (Ake et al, 1973) as:
\begin{equation}
    Alignment(B_i , B_j) = \frac{1}{N} ||_i^T V_j ||^2_F= \frac{1}{N} \sum_{n=1}^N X_n =  \frac{1}{N} \sum_{n=1}^N \cos^2(\theta_n)
\end{equation}
The alignment score is 0 for orthogonal subspaces (all angles are 90°) and 1 for perfectly aligned subspaces (all angles are 0°). A rough interpretation of in-between scores can be made by taking their arc-cosine. 
Overall, scores below 0.6 are likely to indicate that the two spaces are not aligned.

We observed a small alignment between the semantic and syntactic subspace of Llama models (max: 0.042, min:0.038)(Fig.~\ref{fig:sup_circuit}, in appendix, panel A), with a relatively constant value across layers.
These results indicate that semantic and syntactic subspaces are mostly unaligned and do not recruit a similar pattern of units in the model.

We next addressed the question of the role of each network unit. We measured for each model unit the norm of its probe vector.
(Fig.~\ref{fig:sup_circuit}, in appendix, panel B)
After plotting the distribution of these norms, we observed that probes weights were spread across all units of the layer. 
This is expected as, following previous work \citep{hewitt_structural_2019}, we do not explicitly regularize the norm of the probe, which could have favored the discovery of a sparser solutio
We also observed more outliers for the syntactic subspace than the semantic subspace.
Some units  (40 to 90, depending on the layer) had outlier norm, ranging from 2 to 3 times the median norm.
Some units (max 20) were both outliers for the syntactic and semantic subspace.
We then repeated these measurements across syntactic and semantic-coding layers of the model (Fig.~\ref{fig:sup_circuit}, in appendix, panel C), which revealed that an increasing number of units were both outliers for the syntactic and semantic subspace.

These results suggest that the representation of syntactic and semantic subspace is widely distributed across the network units, with a few sets of units being more strongly read out by the probe, although their contribution is mostly washed out by the influence of the majority.
Consequently, these subspaces form distributed and distinct, although low-dimensional, patterns of activity in the initial activity space.

To better understand the activity of each individual units, and how we could understand them in terms of syntactic and semantic subspace, we then turned to an encoding approach
Indeed, the previous analysis on the optimized probe is not entirely convincing, as the probe is not regularized, such that it could give a high norm to two units while making their contribution cancel out.
To deal with that issue, we tried to predict the model unit response from the activity of the syntactic and semantic subspace.
Indeed, if some units' activity can be predicted and explained by linear combination of the subspace activity, but not by simpler features, they are likely strongly participating in the syntactic or semantic code.

To remove contamination due to training and testing on particular datasets, we explore a novel set of sentences taken from the Podcast ECOG dataset \cite{zada_podcast_2025}. 
This dataset is essentially a transcription of a 30-minute long podcast composed of 5305 words. 
We gathered the activations of Llama2 by inputing block of 100 words from the podcast transcript in the model and gathering the average activity of all tokens associated with each individual words.
We then used the semantic and syntactic probed previously estimates to project each layer activity into a semantic and syntactic subspace. 
For this analysis, we only kept the 34 layers out of the 40 layers for which the semantic and syntactic scores were above 80\% of their respective maximal scores across layers. 
This was done to make sure the subspace instantiated true semantic and syntactic structure rather than another projection of the model activity.

We then ran an independent ridge regression with 5 outer and 5 inner cross-validation folds (sequential fold block), from the subspace activity to each individual unit response.
We then partitioned the variance of each regression into the variance explained by the semantic and syntactic subspace $R2 = R2_{semantic}+R2_{syntax}$.
Remarkably, a small number of units (36 in layer 15) had variance strongly predicted by the activity of the syntactic subspace ($R2_{syntax}>60\%$).
Interestingly, these units matched with the outlier units found in the previous analysis (Fig.~\ref{fig:sup_circuit}, in appendix, panel D). 

We then controlled this regression for the effect of known univariate syntactic features.
For that purpose, we used the 96 univariate “syntactic features” available with the Podcast dataset, which are made as one-hot encoding of part-of-speech and dependency relationship categories. 
These features are supposed to encode syntactic properties while not capturing the distance of the syntactic tree.
We ran two regressions, one with only the univariate features as predictors, and the second with the univariate features and the subspace activity as predictors. 
(Fig.~\ref{fig:sup_circuit}, in appendix, panel E)
Remarkably, adding the subspace activity strongly increased the variance explained, indicating that syntactic features were not sufficient to explain these units' responses.
Crucially, this was the case in outlier units, whose variance was consequently explained by the syntactic tree-coding subspace instead of univariate semantic features.
We additionally verified by eye that these units had no outlier response, computing their mean response across each part-of-speech or dependency category; no clear explanation of their tuning could be found in this manner.
In addition, these units' responses were not strongly correlated across the words of the dataset (mean correlation: 0.002, max across two units: 0.39, standard deviation: 0.125), stressing that they had distinct roles in the encoding of syntactic information.
We repeated this analysis across all syntax-coding layers of the model (Fig.~\ref{fig:sup_circuit}, in appendix, panel F), finding a similar conclusion.

Unlike for syntax, no such effect was observed for the semantic subspace: this subspace of activity was not sufficient by itself to predict a large variance of any single unit response.

Together, these results suggest that the instantiation of semantic structure is a distributed process, spread across the units of the model at each layer. The instantiation of syntactic structure is also a distributed process, but with a few units specializing in syntax-coding.

\begin{figure}[h]
  \centering
  \includegraphics[width=0.99\textwidth]{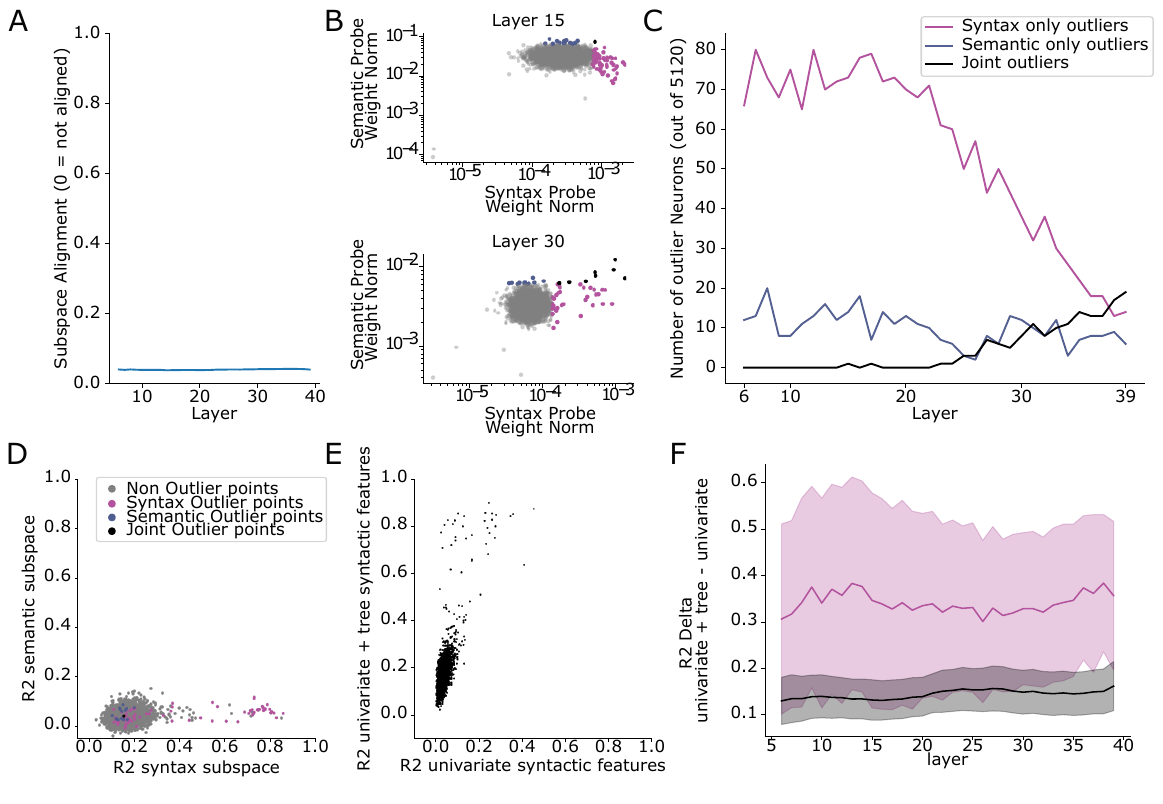}
    \caption{ Subspace alignment and single unit analyses. 
        A: Subspace alignment between the syntactic and semantic subspace across layers instantiating syntax and semantic structure of a Llama2-13B model.
        B: Scatter plot of syntax and semantic probe norm for all units of layer 15 (top) and layer 30 (bottom) of a Llama2-13B model. 
        Units with outlier probe weight norm are highlighted in purple (syntax), navy (semantic) and black (joint outliers).
        C: Number of outlier units for syntax, semantics, and joint outliers across layers instantiating syntax and semantic structure of a Llama2-13B model.
        D: Encoding R2 score of each unit of layer 15 of a Llama2-13B model from the syntactic subspace activity and semantic subspace activity (R2 partitioning).
        To show that outlier units of the structural probe were indeed explained by the syntactic subspace activity, these units' markers are colored in purple.
        E: Unique contribution of the syntactic subspace activity compared to simpler univariate syntactic features for each unit of layer 15 of a Llama2-13B model.
        F: Difference of R2 score (subspace + univariate features - univariate features) across layers instantiating syntax of a Llama2-13B model.
    }
    \label{fig:sup_circuit}
\end{figure}

\subsection*{Probing topology instead of distances}

To complement our distance-based probe, we designed a novel contrastive probe that optimizes the extraction of the topology of the structure, irrespective of the distance.
This probe is inspired by contrastive learning approaches of embedding on WordNet \citep{nickel_poincare_2017}, and more generally by contrastive learning objectives used in word-embedding vectors.

\subparagraph*{Contrastive probe objective.}

The contrastive probe optimizes the contrastive objective of \citep{nickel_poincare_2017}, where we replaced the embeddings with a linear projection from the model activity, and optimize the projection.
More precisely, given a node $i$ from a graph $S$, we define the set of positive nodes $P(i)$ as the nodes directly connected to this node.
We define set of negative nodes $N(i)$ as all other nodes.
We then optimize the following objective:

\begin{equation}
\hat{B} = argmin_{B\in \mathbb{R}^{k,p}} \sum_{i \in S} - \log \left( \frac{\sum_{j \in P(i)} \exp(-||(h_i-h_j)B||_2^2)}{\sum_{k \in N(i)} \exp(-||(h_i-h_k)B||_2^2)} \right)
\end{equation}

In practice, to compute this objective efficiently, we use batched gradient descent, sampling a positive node and a set of negative node.
All hyperparameters searches are performed identically to the distance-based probe, with the exception of the number of epochs, which was the same for syntactic probes, but increased to 1000 epochs for semantic probes.
Indeed, we observed a slower convergence of the semantic probe with the contrastive objective in term of the number of epochs, but since the computational cost of each epoch is low, we could afford this increase in the number of optimization epochs.
The contrastive probe is indeed more memory efficient than the distance-based probe, as it does not require to compute and store all pairwise distances between nodes.

\subparagraph*{Evaluations}

Unlike distance-based probe, the contrastive probe does not optimize the distances between all pairs of nodes.
Consequently, it is expected that distance-based evaluations, like spearman correlation show lower scores for this probe.
Instead, it should reach higher score on topological evaluations, we use the Unlabeled Undirected Attachment Score (UUAS) for syntax, and rank-based evaluations for syntax and semantics.

\subparagraph*{Results}

Results obtained with the contrastive probe confirmed the results obtained with the distance-based probe.
The topology of the semantic WordNet graph emerged in middle-late layers of both text and audio models (Fig.~\ref{fig:sup_contrastive} in appendix, panel A).
English-exposed model showed better instantiation of this topology than music-exposed, environmental sounds-exposed or French-exposed control models (Fig.~\ref{fig:sup_contrastive} in appendix, panel B).
And the instantiation of this topology increased with model size (Fig.~\ref{fig:sup_contrastive} in appendix, panel B).
As observed with the distance-based probe, instantiation of semantic topology progressively emerged during training (Fig.~\ref{fig:sup_contrastive} in appendix, panel C), and saturated for the audio model.
Similar results were obtained for the syntactic tree topology (Fig.~\ref{fig:sup_contrastive} in appendix, panels D to I), where we evaluated the instantiation of the syntactic structure through UUAS and rank scores.
Distance-based probe and contrastive probe led to similar qualitative conclusions although they offer difference quantitative scores because they optimize different objectives (Fig.~\ref{fig:sup_contrastivevsdistance} in appendix).
The distance-based probe led to higher scores for distance-based evaluations (spearman correlation), while the contrastive probe led to higher scores for topological evaluations (UUAS and rank scores).

Together these results confirm the validity of our measurements using the distance-based probe.

\begin{figure}[h]
  \centering
  \includegraphics[width=0.99\textwidth]{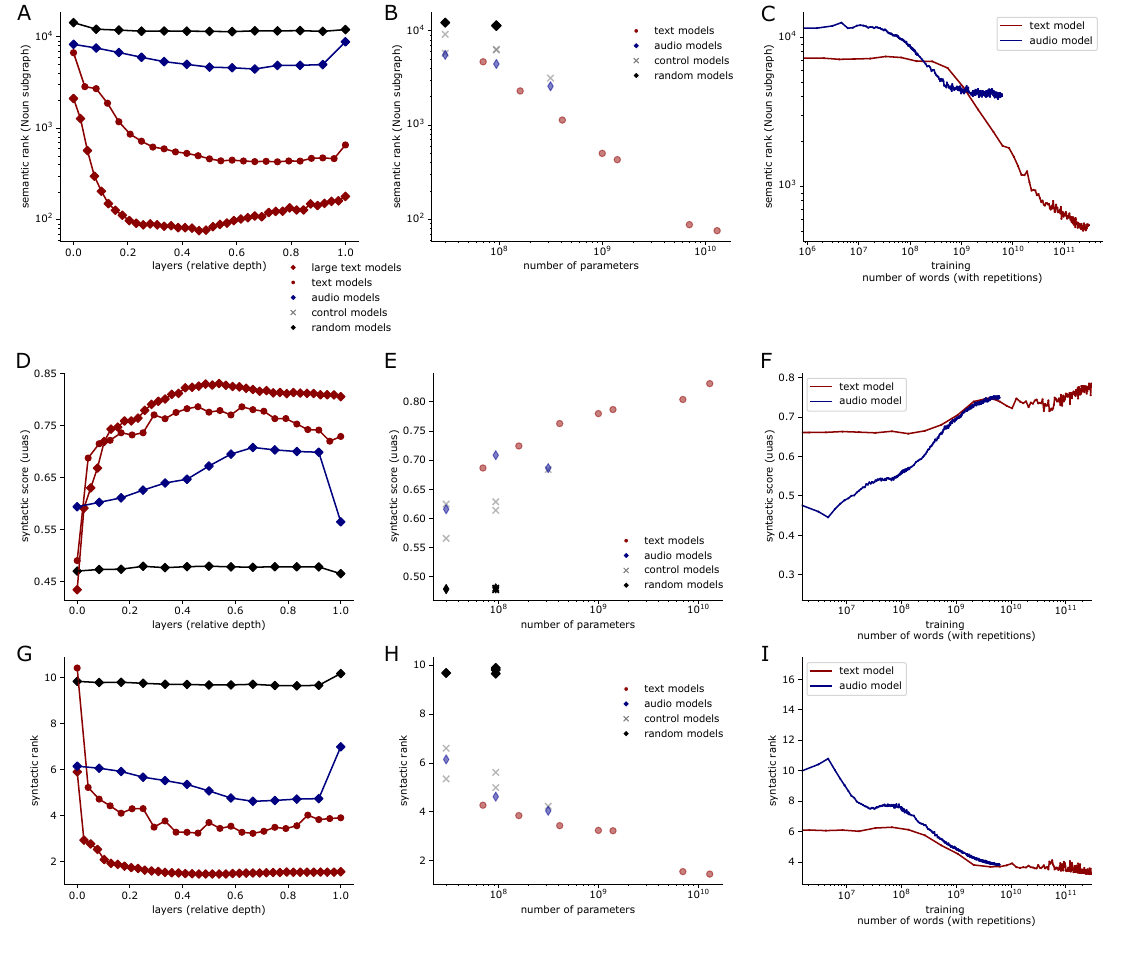}
\caption{A: Semantic rank score of a contrastive probe for a large text (Llama-13B), text (pythia-1.4B), audio (Wav2vec 2.0-94M base), and random (Wav2Vec 2.0-94M base) models.
B: Semantic rank scores of a contrastive probe for all models as a function of the model size.
C: Semantic rank score of a contrastive probe as a function of the quantity of pertaining for the text (pythia-1.4B) and audio (Wav2vec 2.0-94M base) models.
D-F: Same as A-C but for the syntactic dataset evaluated with the UUAS score.
G-I: Same as D-F but for the rank score.
}
\label{fig:sup_contrastive}
\end{figure}

\begin{figure}[h]
  \centering
  \includegraphics[width=0.99\textwidth]{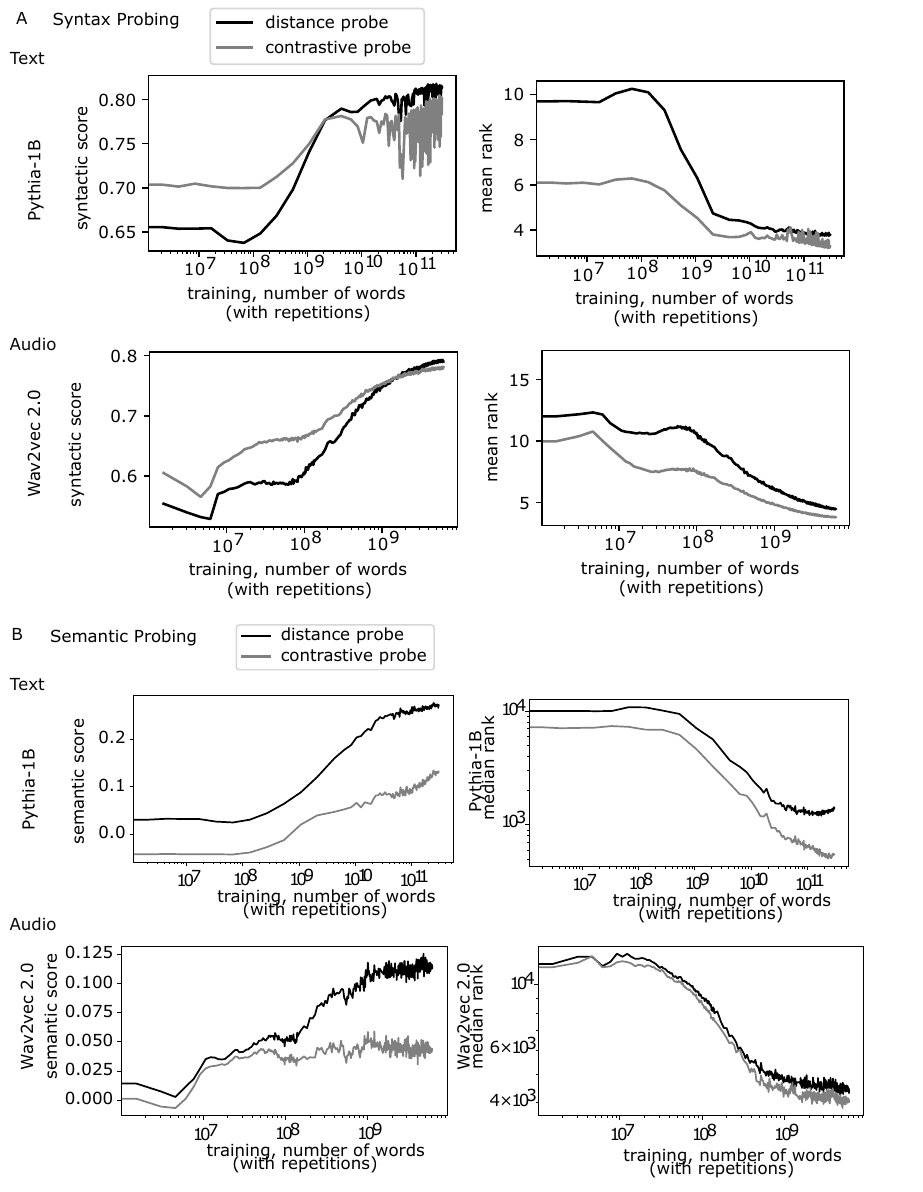}
\caption{A: Score of the structural (distance-based) and contrastive probe on the syntax dataset for a text (pythia-1.0B) and audio (Wav2vec 2.0-94M base) model. 
From left to right, we report for every checkpoint the spearman correlation (syntactic score) and mean rank score.
B: Same as A, but for the semantic dataset, reporting spearman correlation (semantic score) and median rank score.
}
\label{fig:sup_contrastivevsdistance}
\end{figure}

\end{document}